\documentclass{article}

\usepackage{microtype}
\usepackage{graphicx}
\usepackage{booktabs} 

\usepackage{hyperref}

\usepackage[textsize=tiny]{todonotes}

\usepackage{amsmath}
\usepackage{amssymb}
\usepackage{mathtools}
\usepackage{amsthm}

\usepackage{bm}
\usepackage{xcolor}
\usepackage{color-edits}
\usepackage{array}
\usepackage[inline]{enumitem}

\usepackage{subfig}
\usepackage{multirow}

\usepackage{adjustbox}
\usepackage{lipsum}

\usepackage{wrapfig}  
\usepackage{caption}  

\usepackage[conf={restate,no link to proof}]{proof-at-the-end}   

\usepackage{multicol}   

\theoremstyle{plain}
\newtheorem{theorem}{Theorem}[section]

\newtheorem{example}[theorem]{Example}

\theoremstyle{definition}
\newtheorem{definition}[theorem]{Definition}

\theoremstyle{remark}

\usepackage{xspace}     
\newcommand{\ftrl}{\textsc{DP-BandMF}\xspace}
\newcommand{\dpsgd}{\textsc{DP-SGD}\xspace}
\newcommand{\clipnorm}{\ensuremath{\zeta}}
\newcommand{\dataset}{\ensuremath{\mathcal{D}}}

\newcommand{\techname}{NoiseCurve\xspace}

\usepackage[capitalize,noabbrev]{cleveref}

\crefname{problem}{Problem}{Problems}
\Crefname{problem}{Problem}{Problems}

\crefname{claim}{claim}{claims}
\Crefname{claim}{Claim}{Claims}

\crefname{assumption}{assumption}{assumptions}
\Crefname{assumption}{Assumption}{Assumptions}

\usepackage{chngcntr} 

\newcounter{problem}

\newenvironment{problem}
  {%
    \refstepcounter{problem}
    \begingroup
      \begin{equation}
  }
  {%
      \end{equation}
    \endgroup
  }

\usepackage{amsmath,amsfonts,bm}

\def\eqref#1{equation~\ref{#1}}

\def\1{\bm{1}}

\def\vb{{\bm{b}}}

\def\vd{{\bm{d}}}

\def\vg{{\bm{g}}}

\def\vu{{\bm{u}}}
\def\vv{{\bm{v}}}
\def\vw{{\bm{w}}}
\def\vx{{\bm{x}}}
\def\vy{{\bm{y}}}
\def\vz{{\bm{z}}}

\def\mC{{\bm{C}}}
\def\mD{{\bm{D}}}

\def\mH{{\bm{H}}}
\def\mI{{\bm{I}}}

\def\mM{{\bm{M}}}

\def\mT{{\bm{T}}}

\def\mV{{\bm{V}}}
\def\mW{{\bm{W}}}
\def\mX{{\bm{X}}}

\def\mZ{{\bm{Z}}}

\DeclareMathAlphabet{\mathsfit}{\encodingdefault}{\sfdefault}{m}{sl}
\SetMathAlphabet{\mathsfit}{bold}{\encodingdefault}{\sfdefault}{bx}{n}

\newcommand{\E}{\mathbb{E}}

\newcommand{\R}{\mathbb{R}}

\DeclareMathOperator*{\argmin}{arg\,min}

\DeclareMathOperator{\Tr}{Tr}

\def\notationcolor{black} 

\newcommand{\notation}[2]{\newcommand{#1}{{\textcolor{\notationcolor}{\ensuremath{#2}}}}}

\notation{\model}{\Phi}
\notation{\w}{\vw} 
\notation{\pw}{\widetilde{\w}} 
\notation{\z}{\vz} 
\notation{\T}{T} 
\notation{\clip}{c} 
\notation{\I}{\mI}
\notation{\loss}{\mathcal{L}}
\notation{\trace}{\text{trace}}
\notation{\mm}{\mathfrak{\mT}}  
\notation{\vtw}{\hat{\vw}}  
\notation{\cz}{\widetilde{\vz}} 
\notation{\vlambda}{\bm{\lambda}} 
\notation{\transpose}{\top} %
\notation{\upperH}{\hat{\mH}}
\notation{\upperHapprox}{\tilde{\mH}}

\newcommand{\ind}{\mathrel{\perp\mspace{-10mu}\perp}}

\newcommand{\xdownarrow}[1]{%
  {\left\downarrow\vbox to #1{}\right.\kern-\nulldelimiterspace}
}

\newcommand{\xuparrow}[1]{%
  {\left\uparrow\vbox to #1{}\right.\kern-\nulldelimiterspace}
}

\let\grad\nabla

\newcommand{\diag}{\ensuremath{\texttt{diag}}}

\newcommand{\N}{\ensuremath{\mathcal{N}}}
\newcommand{\Mu}{\ensuremath{\mM}}
\newcommand{\one}{\ensuremath{\mathbf{1}}}
\newcommand{\zero}{\ensuremath{\mathbf{0}}}

\usepackage[preprint]{icml2026/icml2026}

\icmltitlerunning{Correlating Cross-Iteration Noise for DP-SGD using Model Curvature}
\addauthor{xin}{blue}
\addauthor{todo}{red}
\addauthor{status}{gray}

\begin{document}

\twocolumn[
  \icmltitle{Correlating Cross-Iteration Noise for DP-SGD using Model Curvature}



  \icmlsetsymbol{equal}{*}

  \begin{icmlauthorlist}
    \icmlauthor{Xin Gu}{psu}
    \icmlauthor{Yingtai Xiao}{tt}
    \icmlauthor{Guanlin He}{psu}
    \icmlauthor{Jiamu Bai}{psu}
    \icmlauthor{Daniel Kifer}{psu}
    \icmlauthor{Kiwan Maeng}{psu}
  \end{icmlauthorlist}

  \icmlaffiliation{psu}{Penn State University}
  \icmlaffiliation{tt}{TikTok Inc}

  \icmlcorrespondingauthor{Xin Gu}{xingu@psu.edu}

  \icmlkeywords{Machine Learning, ICML}

  \vskip 0.3in
]



\printAffiliationsAndNotice{}  

\begin{abstract}
  Differentially private stochastic gradient descent (DP-SGD) offers the promise of training deep learning models while mitigating many privacy risks. However, there is a large accuracy gap between DP-SGD and normal SGD training. 
  One popular line of work to reduce such accuracy gap, known as DP-MF, adds privacy noise correlated across training iterations, so that later noise partially cancels out earlier noise. In this paper, we build upon a crucial observation that the noise affects the training trajectory in two ways --- the noise alters the gradient itself, and the altered gradient changes where the subsequent gradient is computed --- but prior DP-MF works only address the former.
  We propose a technique, \techname, that uses model curvature information estimated from public unlabeled data to address both sources of errors.
  %
  Our experiments on various computer vision and NLP tasks show that \techname consistently improves accuracy over the state-of-the-art DP-MF scheme, \ftrl.
\end{abstract}

\section{Introduction}\label{sec:intro}

Differential privacy (DP;~\citet{dp}) is a rigorous mathematical framework that limits the amount of personal information an attacker can infer from the output of an algorithm that processes confidential data. Differentially private stochastic gradient descent (\dpsgd;~\citet{dpsgd}) is one of the most popular methods for training machine learning (ML) models with DP guarantees. \dpsgd differs from standard SGD in two important ways. First, the gradient of each sample within a batch is separately calculated and clipped by a constant $\clipnorm$ before being averaged. Second, \textit{independent Gaussian noise} is added to the averaged (clipped) gradient before model weights are updated. Although promising, the barrier to wide-spread adoption is the accuracy gap between models trained with \dpsgd and models trained without this privacy protection.  


\begin{figure}[t]
  \centering
  \vspace{-5pt}
  \includegraphics[width=0.35\textwidth]{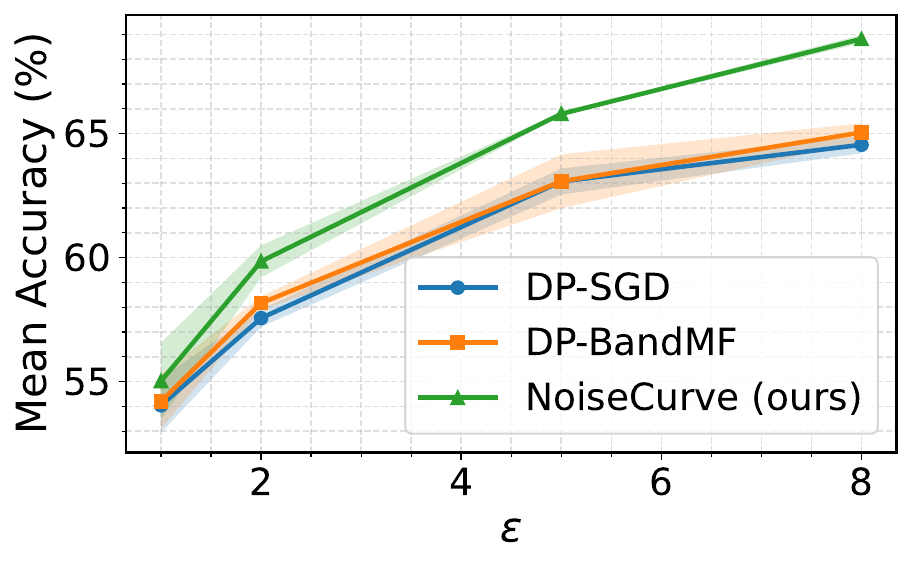}
  \vspace{-10pt}
  \caption{Test accuracy (averaged over 6 runs) on CIFAR-10+VGG, with varying $\epsilon$, $\delta=10^{-5}$.
  $\upperH$ estimated with TinyImageNet.
  Non-private accuracy is 84.05\%.
  Details in \Cref{sec:experiments}.
  }
  \vspace{-10pt}
  \label{fig:vgg}
\end{figure}

To improve the accuracy of \dpsgd, several recent works, which we call \emph{DP-MF} methods, replaced independent Gaussian noise with noise that is  \textbf{correlated across iterations}. The correlations are carefully designed so that noise from some iterations can partially cancel out noise from other iterations, while providing the same privacy~\citep{dpftrl,  bandmf}. The noise correlation is controlled by a mixing matrix $\mC$, which is calculated through minimizing a heuristic objective function subject to privacy constraints. 
The objective function serves as a data-independent approximation of how the noise would cause parameter updates in \dpsgd to differ from ordinary SGD.

We make the critical observation that the noise in each iteration causes the training trajectory of \dpsgd to diverge from SGD in \emph{two} important ways: (1) as is widely recognized, it makes the gradient noisy; (2) as a result, it alters the point at which the gradient of the \emph{subsequent} iteration is computed --- the change in this subsequent gradient is affected by the Hessian of the loss function (details in \Cref{sec:motivation}). Existing DP-MF objective functions only consider the first error and \emph{disregard the second}, most likely because the second error appears to be extremely data-dependent and hard to approximate.
However, we found that an upper bound of the Hessian can be obtained from unlabeled public datasets, and this information can be used to construct a better mixing matrix $\mC$ that significantly improves accuracy over the state-of-the-art (see Figure \ref{fig:vgg}).



Based on a mathematical analysis, 
we propose a new objective function for DP-MF noise correlations that better accounts for the two sources of errors in DP-SGD. The new objective function only needs the eigenvalues of a Hessian upper bound, which can be estimated from unlabeled public dataset.
%
Our proposed approach, \emph{\techname}, is a drop-in replacement for the objective function in existing DP-MF frameworks. 
Below, we summarize our contributions:
\begin{enumerate}[noitemsep, leftmargin=*, topsep=0pt]
    \item \textbf{We propose a new objective that accounts for the interaction between model curvature and DP-SGD with correlated noise.}
    The new objective uses the eigenvalues of a Hessian upper bound, and is a drop-in replacement for
    existing DP-MF software packages. 
    \item \textbf{We show how to estimate the eigenvalues of a Hessian upper bound from unlabeled public data,} with theoretical and empirical justifications. We extend the approach to larger models through a compute-efficient approximation, called curve fitting.
    
    \item\textbf{\techname consistently outperforms the state-of-the-art (SOTA)} DP-MF method, \ftrl, across all the privacy budgets, datasets, and models we tested, showing an accuracy improvement up to 4\% (\Cref{fig:vgg}).

\end{enumerate}

\section{Related Work}   \label{sec:related}
\paragraph{Correlated Noise and DP-MF.}
Correlated noise in DP training was first introduced by~\citet{dpftrl} using tree aggregation; later work (DP-MF family) used matrix factorization instead~\citep{mf-dp-ftrl}. \citet{multiepoch-mf} extended it to multi-epoch training, and \ftrl~\citep{bandmf} enabled privacy amplification.
%
Other improvements include: extending DP-MF to adaptive optimizers~\citep{dpmf-adaptive}, optimizing matrix factorization error~\citep{inv-band}, and leveraging special matrix structures for efficiency~\citep{blt1, blt22, mf-book}.
A concurrent work~\citep{dpmf+} used theoretical convergence analysis to propose rescaling different parts of their objective function. 
To the best of our knowledge, we are the first to improve the correlated noise using the approximated model curvature (Hessian upper bound).

%

\vspace{-10pt}
\paragraph{Using Public Data.}
Pretraining with public data and finetuning with DP can improve utility in DP training
~\citep{YuNBGIKKLMWYZ22, LiTLH22,unlocking, pretrain1, pretrain2}. However, achieving a reasonable accuracy requires significant cost and a large public dataset. 
Public data can also be used to pre-condition the noisy \dpsgd gradients \cite{bypass,donot}.
These techniques are orthogonal to, and can be used in conjunction with, our approach. 
Also, \techname does not require \emph{labeled} public data, unlike these approaches.
%

\vspace{-10pt}
\paragraph{Faster Convergence.}
\dpsgd can be seen as a composition of individual DP mechanisms applied to each weight update, with the total privacy cost growing sublinearly of the total number of training iterations. Therefore, faster convergence results in lower privacy costs or, equivalently, lower noise scale under the same privacy budget. This line of work includes: using information from noisy history~\citep{dp2}, integrating second-order information~\citep{second-order-dp}, adaptively determining privacy parameters~\citep{adapt-budget, adapt-clip}. Again, these improvements are orthogonal to our work.
%


\section{Motivation}\label{sec:motivation}

Given a model $\model$ with learnable weights $\w\in\R^p$ and loss function $\loss$, let $\w_0, \w_1, \w_2, \dots, \w_\T$ be the weights after each update of SGD. $\w_0$ are the initial weights and $\w_{i+1}=\w_i - \eta\grad \loss(\w_i)$, where $\eta$ is the learning rate. To simplify the discussion for this section, we assume all gradients have norm smaller than some constant $\clipnorm$ (e.g., the gradients are clipped). In each iteration $i$, DP-SGD adds isotropic Gaussian noise $\z_i\sim N(\zero, \sigma^2\I_p)$ to each (clipped) gradient, resulting in the weight sequence $\pw_0, \pw_1,\pw_2, \dots, \pw_\T$, where $\pw_0=\w_0$ and $\pw_{i+1}=\pw_i - \eta(\grad \loss(\pw_i) + \z_i)$. Each Gaussian noise is independent: $\z_i\ind \z_j$ when $j\neq i$.

The DP-MF
line of work correlates noise across different iterations by searching for a {lower triangular} mixing matrix $\mC\in\R^{\T\times\T}$ {whose columns have norm 1}. It generates noise vectors $\widetilde{\z}_0, \dots \widetilde{\z}_{\T-1}$ from $\z_0,\dots, \z_{\T-1}$ through: $[\widetilde{\z}_0, \dots \widetilde{\z}_{\T-1}] = [\z_0, \z_2, \dots, \z_{\T-1}]\mC^{-\transpose}$. 
That is, each $\widetilde{\z}_i$ is a linear combination of $\z_0, \z_2, \dots, \z_{i}$. 
During training, the new noise vectors are used instead, i.e., $\pw_{i+1}=\pw_i - \eta(\grad \loss(\pw_i) + \widetilde{\z}_i)$.
%
A proper choice of $\mC$ can let DP-MF outperform DP-SGD at the same privacy cost~\citep{mf-dp-ftrl, bandmf}. In the current state-of-the-art methods (e.g., \ftrl), $\mC$ is obtained by solving a data-independent optimization problem \citep{mf-dp-ftrl,bandmf} to
approximately minimize the sum of the squared errors of $\pw_{1},\dots, \pw_{\T}$ (i.e., minimize $\sum_{i=0}^{\T-1} ||\widetilde{\z}_i||_2^2$), subject to privacy constraints. That is, the error of the parameter estimate $\pw_{i}$ is modeled only through $||\widetilde{\z}_{i-1}||_2^2$, the amount of noise in the $(i-1)^\text{th}$ iteration.

However, the difference between $\pw_{i}$ and $\w_i$ depends not just on the noise used in iteration $i-1$; it also depends on how prior noise had altered where the gradients are taken, i.e., $\grad\loss(\pw_{i-1})$ vs. $\grad\loss(\w_{i-1})$. To see this more clearly, consider the case of $i=0$ and $i=1$. Recall that $\w_0$ is the initial value of the weights and is the same for both private and non-private training:
\begin{align*}
    \pw_1 - \w_1 &= (\w_0 - \eta (\grad\loss(\w_0)+\widetilde{\z}_0)) - (\w_0 - \eta \grad\loss(\w_0)) \\
                &= -\eta\widetilde{\z}_0.\\
    \pw_2 - \w_2 &= (\pw_1 - \eta (\grad\loss(\pw_1)+\widetilde{\z}_1)) - (\w_1 - \eta \grad\loss(\w_1))\\
    %
    %
    &=-\eta\widetilde{\z}_0 - \eta\widetilde{\z}_1 - \eta(\grad\loss(\pw_1)-\grad\loss(\w_1))\\
    &\approx -\eta\widetilde{\z}_0  -\eta\widetilde{\z}_1 - \eta\grad^2\loss(\w_1)(\pw_1-\w_1)\\
    &=-\eta (\widetilde{\z}_0 + \widetilde{\z}_1) + \eta^2 \grad^2\loss(\w_1)\widetilde{\z}_0.
\end{align*}
While previous works on DP-MF optimize the first term of the last equation ($-\eta(\widetilde{\z}_0 + \widetilde{\z}_1)$), they largely ignore the affect of the second term ($\eta^2 \grad^2\loss(\w_1)\widetilde{\z}_0$), which is another source of error.
The second term shows that the Hessian of the loss (i.e., model curvature)
plays an important role in determining how the privacy-preserving update path $\w_0, \pw_1, \pw_2, \dots$ diverges from the non-private path $\w_0, \w_1, \w_2, \dots$.
%
In \Cref{sec:obj}, we propose an alternative objective function, which incorporates such an (upper bound of) Hessian. In \Cref{sec:upper_bound}, we discuss how to estimate Hessian upper bounds for realistic deep learning settings.

\section{A New Objective for Correlated Noise}\label{sec:obj}
We first perform an exact DP-MF convergence analysis for quadratic loss with known Hessian  $\mH$.
Then, under the assumption there is an upper bound $\upperH$ on the Hessian, we extend the analysis to non-convex loss functions (proofs in \Cref{app:proofs}). We use these results to identify a new objective for the mixing matrix $\mC$.
In Section~\ref{sec:application_to_dp_mf}, we explain how to use it in the DP-MF framework. 

\subsection{Convergence Analysis}
\label{sec:quad}

First, we analyze the quadratic loss with known Hessian $\mH$.
%
Let $\loss(\vw)= \frac{1}{2}(\vw-\vd)^\transpose \mH (\vw-\vd)$ be the loss function for some fixed vector $\vd$ and positive semi-definite Hessian $\mH\in\R^{p\times p}$.
Note that $\grad \loss(\vw)=\mH(\vw-\vd)$ and $\grad^2\loss(\vw)=\mH$. The noise-free update path, starting from initial weights $\vw_0$ is $\vw_0, \vw_1, \dots, \vw_\T$ where $\vw_i=\vw_{i-1} - \eta\grad\loss(\vw_{i-1})=\vw_{i-1}-\eta \mH(\vw_{i-1}-\vd)$. 
Given i.i.d. vectors $\vz_0, \dots, \vz_{\T-1}$ with $\vz_i\sim N(0, \sigma^2\I_p)$, the correlated noise vectors are $[\widetilde{\vz}_0, \dots, \widetilde{\vz}_{\T-1}] = [\vz_0, \dots, \vz_{\T-1}]\mC^{-\transpose}$.
The correlated-noise update path, starting from initial weights $\pw_0=\vw_0$ is $\pw_0, \pw_1, \dots, \pw_{\T}$ where $\pw_i = \pw_{i-1}-\eta(\mH(\pw_{i-1}-\vd)+\widetilde{\vz}_{i-1})$.
The following theorem computes both (1) the expected difference in loss between the last noise-free iteration $\loss(\vw_\T)$ and the last correlated-noise iteration $\loss(\pw_\T)$ and (2) the traditional convergence criterion studied in non-convex analysis of DP-SGD: $\frac{1}{T}\sum_{i=1}^\T||\grad\loss(\pw_i)||_2^2$.

\begin{theoremEnd}[proof end,category=quadratic]{theorem}\label{thm:risk}
    Under quadratic loss, define $\mX=\mC^\transpose\mC$. Let $\mu_0, \dots, \mu_{p-1}$ be the eigenvalues (including multiplicity) of $\mH$. Define the diagonal matrix $\mM=\text{Diag}(\mu_0,\dots, \mu_{p-1})$ and the matrix $\mV\in \R^{p\times \T}$ so that $\mV[i, j]=(1-\eta\mu_i)^{\T - j - 1}$ (indexing starts at 0). Then, the difference in expected loss between gradient descent with correlated noise and noise-free gradient descent is:
    \begin{align}
    &\hspace{-0.5em}E[\loss(\pw_\T)]\hspace{-0.2em}-\hspace{-0.2em}E[\loss(\vw_\T)] = \frac{\eta^2\sigma^2}{2}\Tr(\mX^{-1}(\mV^\transpose\mM\mV))\label{eqn:thm:q1}\\
    &\hspace{-0.5em}\sum\limits_{i=1}^\T\hspace{-0.2em}\frac{||\grad\loss(\pw_i)||_2^2}{T}\hspace{-0.2em}=\hspace{-0.2em}k^* \hspace{-0.5em}+\hspace{-0.2em}\eta^2\sigma^2 \Tr\left(\mX^{-1} (\mm\hspace{-0.2em}\odot\hspace{-0.2em}\mV^{\top}\hspace{-0.2em}\Mu^2 \mV) \right)\label{eqn:thm:q2}
    \end{align}
    where $k^*$ is a constant independent of the noise correlation, and $\mm[i,j]=(\T-\max(i,j))/\T$. That is,
    \begin{align*}
    \mm = \begin{pmatrix}
            T & T -1  & \cdots & 1 \\
            T-1 & T-1 & \cdots & 1 \\
            \vdots          & \vdots           & \ddots  & \vdots \\
            1 & 1  & \cdots & 1
     \end{pmatrix}/\T
    \end{align*}
\end{theoremEnd}
\begin{proofEnd}
    The first step is to obtain closed-form expressions for  $\pw_i-\vw_i$, for each $i$, in terms of the noise vectors, Hessian, and initialization. We observe that
    \begin{align}
        \pw_1 - \vw_1 &= (\vw_{0}-\eta (\mH(\vw_{0}-\vd)+\widetilde{\vz}_0)) - (\vw_{0}-\eta \mH(\vw_{0}-\vd))=-\eta\widetilde{\vz}_0\nonumber\\
        \pw_2 - \vw_2 &= (\pw_{1}-\eta (\mH(\pw_{1}-\vd)+\widetilde{\vz}_1)) - (\vw_{1}-\eta \mH(\vw_{1}-\vd))\nonumber\\
        &= (\pw_{1} - \vw_{1}) - \eta\mH(\pw_1-\vw_1) -\eta \widetilde{\vz}_1\nonumber\\
        &= (\mI - \eta\mH) (\pw_{1} - \vw_{1}) -\eta \widetilde{\vz}_1
        = -\eta(\mI - \eta\mH) \widetilde{\vz}_0 -\eta \widetilde{\vz}_1\nonumber\\
        \pw_3 - \vw_3 &=(\mI - \eta\mH) (\pw_{2} - \vw_{2}) -\eta \widetilde{\vz}_2
        = -\eta(\mI - \eta\mH)^2 \widetilde{\vz}_0 -\eta(\mI - \eta\mH) \widetilde{\vz}_1-\eta \widetilde{\vz}_2\nonumber\\
        \pw_i-\vw_i &= -\eta\sum\limits_{j=0}^{i-1} (\mI - \eta\mH)^j\widetilde{\vz}_{i-j-1}\label{eqn:quaddiff}
    \end{align}
    Then we calculate the difference in losses for the final noisy and non-noisy parameters:
    \begin{align*}
        \loss(\pw_\T) - \loss(\vw_\T) &= \frac{1}{2} (\pw_\T-\vd)^T\mH(\pw_\T-\vd) - \frac{1}{2} (\vw_\T-\vd)^T\mH(\vw_\T-\vd)\\
        &=\frac{1}{2} \left(\pw_\T-\vw_\T +(\vw_\T -\vd)\right)^T\mH\left(\pw_\T-\vw_\T +(\vw_\T -\vd)\right) - \frac{1}{2} (\vw_\T-\vd)^T\mH(\vw_\T-\vd)\\
        &=\frac{1}{2}(\pw_\T-\vw_\T)^T\mH(\pw_\T-\vw_\T) +  (\pw_\T-\vw_\T)^T\mH(\vw_\T -\vd)\\
        E[\loss(\pw_\T) - \loss(\vw_\T)] &= E\left[\frac{1}{2}(\pw_\T-\vw_\T)^T\mH(\pw_\T-\vw_\T)\right]
        + E\left[\left(-\eta\sum\limits_{j=0}^{\T-1} (\mI - \eta\mH)^j\widetilde{\vz}_{\T-j-1}\right)^T\mH(\vw_\T -\vd)\right]\\
        &=E\left[\frac{1}{2}\left(-\eta\sum\limits_{j=0}^{\T-1} (\mI - \eta\mH)^j\widetilde{\vz}_{\T-j-1}\right)^T\mH\left(-\eta\sum\limits_{j=0}^{\T-1} (\mI - \eta\mH)^j\widetilde{\vz}_{\T-j-1}\right)\right]\\
        &\hspace{-3cm}=E\left[\frac{1}{2}\left(-\eta\sum\limits_{j=0}^{\T-1} (\mI - \eta\mH)^j(\sum_i C^{-1}[\T-j-1, i]\vz_i)\right)^T\mH\left(-\eta\sum\limits_{j=0}^{\T-1} (\mI - \eta\mH)^j(\sum_i C^{-1}[\T-j-1, i]\vz_i)\right)\right]\\
        &\hspace{-2cm}=\frac{\eta^2}{2}\sum\limits_{j=0}^{\T-1}\sum\limits_{\ell=0}^{\T-1}\sum_i E\left[\left( (\mI - \eta\mH)^j C^{-1}[\T-j-1, i]\vz_i\right)^T\mH\left( (\mI - \eta\mH)^\ell C^{-1}[\T-\ell-1, i]\vz_i\right)\right]\\
        &\hspace{-2cm}=\frac{\eta^2}{2}\sum\limits_{j=0}^{\T-1}\sum\limits_{\ell=0}^{\T-1}\sum_i E\left[\left( C^{-1}[\T-j-1, i]\vz_i\right)^T\mH(\mI - \eta\mH)^{j+\ell} \left( C^{-1}[\T-\ell-1, i]\vz_i\right)\right]\\
        &=\frac{\eta^2}{2}\sum\limits_{j=0}^{\T-1}\sum\limits_{\ell=0}^{\T-1}\sum_i  C^{-1}[\T-j-1, i]C^{-1}[\T-\ell-1, i]\trace(\mH(\mI - \eta\mH)^{j+\ell}) \\
        \intertext{Now define $\mX=\mC^T\mC$}
        &=\sigma^2\frac{\eta^2}{2}\sum\limits_{j=0}^{\T-1}\sum\limits_{\ell=0}^{\T-1}\mX^{-1}[\T-j-1, \T-\ell-1]\trace(\mH(\mI - \eta\mH)^{j+\ell}) \\
        &=\sigma^2\frac{\eta^2}{2}\sum\limits_{j=0}^{\T-1}\sum\limits_{\ell=0}^{\T-1}\mX^{-1}[j,\ell]\trace(\mH(\mI - \eta\mH)^{2\T - j-\ell-2}) \\
        \intertext{Let $\mu_0,\dots, \mu_{p-1}$ be the eigenvalues (including multiplicity) of $\mH$ sorted in non-decreasing order}
        &=\frac{\eta^2\sigma^2}{2}\sum\limits_{j=0}^{\T-1}\sum\limits_{\ell=0}^{\T-1}\mX^{-1}[j,\ell]\sum_{i=0}^{p-1}\mu_i(1 - \eta\mu_i)^{\T - j-1}(1 - \eta\mu_i)^{\T -\ell-1} \\
        \intertext{Define the diagonal matrix $\mM=\text{Diag}(\mu_0,\dots, \mu_{p-1})$ and the matrix $\mV\in \R^{p\times \T}$ so that $\mV[i, j]=(1-\eta\mu_i)^{\T - j - 1}$}\\
        &=\frac{\eta^2\sigma^2}{2}\sum\limits_{j=0}^{\T-1}\sum\limits_{\ell=0}^{\T-1}\mX^{-1}[j,\ell](\mV^T\mM\mV)[j,\ell] =\frac{\eta^2\sigma^2}{2}\trace(\mX^{-1}(\mV^T\mM\mV))\\
    \end{align*}    
    \textbf{This proves the first part about parameter drift.}
    
    \textbf{Next to prove the part about the average of the squared gradient norms}, we note that $\grad\loss(\vw_j) = \mH(\vw_j-\vd)$ and $\grad\loss(\pw_j) = \mH(\pw_j-\vd)$ and so $\grad\loss(\pw_j)-\grad\loss(\vw_j) = \mH((\pw_j-\vd)-(\vw_j-\vd))$. From Equation \ref{eqn:quaddiff} we note that $E[((\pw_j-\vd)-(\vw_j-\vd))^t\mH^t\mH(\vw_j-\vd)]=0$ and hence 
    \begin{align}\label{eqn:gradlosssimple}
        E[||\grad\loss(\pw_j)-\grad\loss(\vw_j)||_2^2]&= E\left[((\pw_j-\vd)-(\vw_j-\vd))^t \mH^t\mH((\pw_j-\vd)-(\vw_j-\vd))\right]\nonumber\\
        &=E\left[(\pw_j-\vd)^t \mH^t\mH((\pw_j-\vd)-(\vw_j-\vd))\right]\nonumber\\
        &=E\left[(\pw_j-\vd)^t \mH^t\mH(\pw_j-\vd)\right] - (\vw_j-\vd)^t \mH^t\mH(\vw_j-\vd)\nonumber\\
        &=E[||\grad\loss(\pw_j)||_2^2]-||\grad\loss(\vw_j)||_2^2
    \end{align}
    Combining Equations \ref{eqn:quaddiff} and \ref{eqn:gradlosssimple}, we get:

    \begin{align*}
        \frac{1}{\T}\sum\limits_{i=1}^\T ||\grad\loss(\pw_i)||_2^2 
        &=\frac{1}{\T}\sum\limits_{i=1}^\T E[||\grad\loss(\pw_i)-\grad\loss(\vw_i)||_2^2] + \frac{1}{\T}\sum\limits_{i=1}^\T||\grad\loss(\vw_i)||_2^2\\
        &=\frac{1}{\T}\sum\limits_{i=1}^\T E[||\mH(\pw_i-\vw_i)||_2^2] + \frac{1}{\T}\sum\limits_{i=1}^\T||\grad\loss(\vw_i)||_2^2\\
        &=\frac{1}{\T}\sum\limits_{i=1}^\T E[||\mH(\pw_i-\vw_i)||_2^2] + k^*\\
        \intertext{where $k^*=\frac{1}{\T}\sum\limits_{i=1}^\T||\grad\loss(\vw_i)||_2^2$ is a constant that does not depend on noise}
        &=k^*+\frac{1}{\T}\sum\limits_{i=1}^\T E\left[\left|\left|\eta\sum\limits_{j=0}^{i-1} \mH(\mI - \eta\mH)^j\widetilde{\vz}_{i-j-1}\right|\right|_2^2\right]\\
        &=k^*+\frac{1}{\T}\sum\limits_{i=1}^\T E\left[\left|\left|\eta\sum\limits_{j=0}^{i-1} \mH(\mI - \eta\mH)^j\left(\sum_\ell C^{-1}[i-j-1, \ell]\vz_\ell\right)\right|\right|_2^2\right]
    \end{align*}

    Define
    \begin{align*}
        \vu_i := \eta\sum_{j=0}^{i-1}\mH(\mI - \eta\mH)^j\left(\sum_\ell C^{-1}[i-j-1, \ell]\vz_\ell\right) \in \R^p
    \end{align*}
    Then
    \begin{align*}
        \E||\vu_i||_2^2 
            &= \E[\vu_i^\top \vu_i] \\
            &= \eta^2\sum_{j=0}^{i-1}\sum_{j'=0}^{i-1}\sum_{l}\sum_{l'}C^{-1}[i-j-1, l]C^{-1}[i-j'-1,l']
                \E\left[\vz_l^\top\left(\mH(\mI - \eta\mH)^j\right)^\top\left(\mH(\mI - \eta\mH)^{j'}\right)\vz_{l'}\right] \\
            &= \sigma^2\eta^2\sum_{j=0}^{i-1}\sum_{j'=0}^{i-1}\sum_{l}C^{-1}[i-j-1, l]C^{-1}[i-j'-1,l]
                \Tr\left(\mH^2(\mI - \eta\mH)^{j + j'}\right)   \\
            \intertext{Now define $\mX = \mC^\top \mC$} \\
            &= \sigma^2\eta^2\sum_{j=0}^{i-1}\sum_{j'=0}^{i-1}\mX^{-1}[i-j-1, i-j'-1]
                \Tr\left(\mH^2(\mI - \eta\mH)^{j + j'}\right)   \\
            &= \sigma^2\eta^2\sum_{j=0}^{i-1}\sum_{j'=0}^{i-1}\mX^{-1}[i-j-1, i-j'-1]
                \Tr\left(\mH^2(\mI - \eta\mH)^{j + j'}\right)   \\
            \intertext{Define the diagonal matrix $\mM=\text{Diag}(\mu_0,\dots, \mu_{p-1})$ and the matrix $\mV^{(i)} \in \R^{p\times i}$ so that $\mV^{(i)}[r, j]=(1-\eta\mu_r)^{i - j - 1}$, $j=0, \dots, i-1$, then} \\
            &= \sigma^2\eta^2 \Tr\left(\mX^{-1}_{[:i-1, :i-1]}(\mV^{(i)\top} \Mu^2 \mV^{(i)})\right)
    \end{align*}
    Plug back into that average:
    \begin{align*}
        \frac{1}{\T}\sum\limits_{i=1}^\T ||\grad\loss(\pw_i)||_2^2 
            &= k^* + \frac{1}{\T}\sum\limits_{i=1}^\T \E||\vu_i||_2^2   \\
            &= k^* + \frac{\eta^2\sigma^2}{\T}\sum\limits_{i=1}^\T \Tr\left(\mX^{-1}_{[:i-1, :i-1]}(\mV^{(i)\top} \Mu^2 \mV^{(i)})\right)    \\
            &= k^* + \eta^2\sigma^2 \Tr\left((\mm \odot \mV^{\top} \Mu^2 \mV) \mX^{-1}\right)
    \end{align*}
    where we define $\mV \in \R^{p\times \T} := \mV^{(T)}$ so that $\mV[i, j]=(1-\eta\mu_i)^{\T - j - 1}$, and $\mm[i,j]=(\T-\max(i,j))/\T$
    \begin{align*}
        \mm = \begin{pmatrix}
            T & T -1  & \cdots & 1 \\
            T-1 & T-1 & \cdots & 1 \\
            \vdots          & \vdots           & \ddots  & \vdots \\
            1 & 1  & \cdots & 1
     \end{pmatrix}/\T
    \end{align*}
    That completes the proof.
\end{proofEnd}
Comparing Equations \ref{eqn:thm:q1} and \ref{eqn:thm:q2}, we see that they have almost the same dependency
on the mixing matrix $\mC$ (recall $\mX=\mC^\top\mC$). Equation \ref{eqn:thm:q2} has an $\mM^2$ in place of $\mM$ and also introduces a $\mm$ element-wise product that emphasizes terms related to the earlier iterations of noisy gradient descent. 
The difference between the two equations is due to the fact that they measure different quantities. Equation \ref{eqn:thm:q1} measures the difference in quality (loss), while Equation \ref{eqn:thm:q2} only measures convergence rate. We note that both results are exact for the quadratic case (i.e., they are not upper bounds).

%

Next, we generalize this result to cases where the loss function is non-convex, and only an upper bound $\upperH$ on the Hessians is available (we discuss how to estimate such an $\upperH$ in Section \ref{sec:upper_bound}). The result, Theorem \ref{thm:dpmf-fosp-Hsmooth} (below), matches Equation \ref{eqn:thm:q2} in terms of the dependency on $\mX=\mC^\top\mC$.

Given a function $f$, we say that $\upperH$ is a Hessian upper bound on $f$ if $\upperH-\grad^2f$ is positive semi-definite. Hessian upper bounds have two consequences that we use as assumptions in our convergence analysis:
\begin{itemize}[leftmargin=*, topsep=0pt]
\item \textbf{Upper Bound Property 1}:\\ $||\grad f(\vx)-\grad f(\vy)||_2\leq ||\upperH(\vx-\vy)||_2$.
\item \textbf{Upper Bound Property 2} $\Big($define $||\vv||_{\upperH}=\sqrt{\vv^T \upperH \vv}$$\Big)$:\\ $f(\vy)\leq f(\vx) + \langle f(\vx),\vy-\vx\rangle + \frac{1}{2}||\vy-\vx||^2_\upperH$.
\end{itemize}

\begin{theoremEnd}[proof end,category=nonconvex]{theorem}[Non-convex Convergence]\label{thm:dpmf-fosp-Hsmooth}
     Consider the DP-MF iteration updates:
    \begin{align*}
        \pw_{t} = \pw_{t-1} - \eta \left(\grad \loss(\pw_{t-1}) + \cz_{t-1}\right),
    \end{align*}
    where $\mC$ is an arbitrary lower triangular mixing matrix whose column norms are all $1$, $\mZ \sim \N(0, \sigma^2)^{T \times p}$, and $\cz_{t}$ is the $t$-th row of $\mC^{-1}\mZ$ (indexed from 0). 
    Let  $\loss$ be a non-convex loss function with Hessian upper bound $\upperH$ such that $\loss$ has a minimum value $\loss^*$, let the learning rate $\eta \leq 1/\lambda_{\max}(\upperH)$. Then:
    \begin{align*}
        \frac{1}{T}\sum_{t=0}^{T-1} \|\grad \loss(\pw_t)\|_2^2 &\leq \frac{4\loss(\vw_0) - 4\loss^*}{\eta T} + 10\eta\sigma^2\Tr(\mW\mX^{-1})
    \end{align*}
    where $\mX = \mC^\top\mC$, $\mW = \mm \odot \mV^\top\Mu^2\mV$, and
    $\mm$, $\mM$ and $\mV$ follow the definitions from \Cref{thm:risk}, except they are computed using the eigenvalues of $\upperH$ instead of $\mH$.
\end{theoremEnd}

\begin{proofEnd}
    We consider the following virtual iterates $\{\vv_t\}_{t=0}^T$:
    \begin{align*}
        \vv_{t+1} &= (\mI - \eta\upperH)\vv_t - \eta(\grad\loss(\pw_t) - \upperH\pw_t)  \\
                  &= \vv_t - \eta\left(\grad\loss(\pw_t) + \upperH(\vv_t - \pw_t)\right)
    \end{align*}
    For notation simplicity, let
    \begin{align*}
        \vb_t := \grad\loss(\pw_t) + \upperH(\vv_t - \pw_t)
    \end{align*}
    By the Hessian upper bound $\upperH$ assumption on $\loss$, we have:
    \begin{align*}
        \loss(\vv_{t+1}) 
            &\leq \loss(\vv_t) - \eta\langle\grad\loss(\vv_t), \vb_t\rangle + \frac{\eta^2}{2}\|{\vb_t}\|_{\upperH}^2 \\
            &= \loss(\vv_t) 
                - \frac{\eta}{2}\|\grad\loss(\vv_t)\|_2^2 
                - \frac{\eta}{2}\|\vb_t\|_2^2
                + \frac{\eta}{2}\|\grad\loss(\vv_t) - \vb_t\|_2^2
                + \frac{\eta^2}{2}\|{\vb_t}\|_{\upperH}^2   \\
            &= \loss(\vv_t) 
                - \frac{\eta}{2}\|\grad\loss(\vv_t)\|_2^2 
                + \frac{\eta}{2}\|\grad\loss(\vv_t) - \vb_t\|_2^2
                + \left(-\frac{\eta}{2}\|\vb_t\|_2^2 + \frac{\eta^2}{2}\|{\vb_t}\|_{\upperH}^2\right)
    \end{align*}
    Since $\eta\leq 1/ \lambda_{\max}(\upperH)$, the last term is $\leq 0$ and can be dropped. So,
    \begin{align*}
         \loss(\vv_{t+1}) 
            &\leq \loss(\vv_t) 
                - \frac{\eta}{2}\|\grad\loss(\vv_t)\|_2^2 
                + \frac{\eta}{2}\|\grad\loss(\vv_t) - \vb_t\|_2^2
                + \left(-\frac{\eta}{2}\|\vb_t\|_2^2 + \frac{\eta^2}{2}\|{\vb_t}\|_{\upperH}^2\right)   \\
            &\leq \loss(\vv_t) 
                - \frac{\eta}{2}\|\grad\loss(\vv_t)\|_2^2 
                + \frac{\eta}{2}\|\grad\loss(\vv_t) - \vb_t\|_2^2
    \end{align*}
    Now we work creating a bound on the second term $-\frac{\eta}{2}\|\grad\loss(\vv_t)\|_2^2$:
    \begin{align*}
        \|\grad\loss(\pw_t)\|_2^2
            &= \|\grad\loss(\vv_t) + (\grad\loss(\pw_t) - \grad\loss(\vv_t))\|_2^2  \\
            &\leq 2\|\grad\loss(\vv_t)\|_2^2 + 2\|\grad\loss(\pw_t) - \grad\loss(\vv_t)\|_2^2   \\
        \Rightarrow \|\grad\loss(\vv_t)\|_2^2 &\geq \frac{1}{2}\|\grad\loss(\pw_t)\|_2^2 - \|\grad\loss(\pw_t) - \grad\loss(\vv_t)\|_2^2    \\
        \Rightarrow -\frac{\eta}{2}\|\grad\loss(\vv_t)\|_2^2 &\leq -\frac{\eta}{4}\|\grad\loss(\pw_t)\|_2^2 + \frac{\eta}{2}\|\grad\loss(\pw_t) - \grad\loss(\vv_t)\|_2^2
    \end{align*}
    Substitute back, we have:
    \begin{align*}
        \loss(\vv_{t+1}) 
            &\leq \loss(\vv_t) 
                - \frac{\eta}{2}\|\grad\loss(\vv_t)\|_2^2 
                + \frac{\eta}{2}\|\grad\loss(\vv_t) - \vb_t\|_2^2       \\
            &\leq \loss(\vv_t)
                -\frac{\eta}{4}\|\grad\loss(\pw_t)\|_2^2 + \frac{\eta}{2}\|\grad\loss(\pw_t) - \grad\loss(\vv_t)\|_2^2
                + \frac{\eta}{2}\|\grad\loss(\vv_t) - \vb_t\|_2^2
    \end{align*}
    Using the consequence on $\|\grad\loss(\pw_t) - \grad\loss(\vv_t)\|_2^2$ of the $\upperH$ upper bound, we have:
    \begin{align*}
        \loss(\vv_{t+1}) 
            &\leq \loss(\vv_t)
                -\frac{\eta}{4}\|\grad\loss(\pw_t)\|_2^2 + \frac{\eta}{2}\|\grad\loss(\pw_t) - \grad\loss(\vv_t)\|_2^2
                + \frac{\eta}{2}\|\grad\loss(\vv_t) - \vb_t\|_2^2   \\
            &\leq \loss(\vv_t)
                -\frac{\eta}{4}\|\grad\loss(\pw_t)\|_2^2 
                + \frac{\eta}{2}\|{\upperH(\pw_t - \vv_t)}\|_2^2
                + \frac{\eta}{2}\|\grad\loss(\vv_t) - \vb_t\|_2^2
    \end{align*}
    Now we approach the last term $\|\grad\loss(\vv_t) - \vb_t\|_2^2$, we have:
    \begin{align*}
        \grad\loss(\vv_t) - \vb_t &= \grad\loss(\vv_t) - (\grad\loss(\pw_t) + \upperH(\vv_t - \pw_t))   \\
                                  &= \grad\loss(\vv_t) - \grad\loss(\pw_t) - \upperH(\vv_t - \pw_t)\\
        \|\grad\loss(\vv_t) - \vb_t\|_2^2
            &\leq 2\|\grad\loss(\vv_t) - \grad\loss(\pw_t)\|_2^2 + 2\|\upperH(\vv_t - \pw_t)\|_2^2  \\
            &\leq 4\|\upperH(\vv_t - \pw_t)\|^2\quad\text{(consequence of $\upperH$ upper bound)}
    \end{align*}
    Substitute back, we have:
    \begin{align*}
        \loss(\vv_{t+1}) 
            &\leq \loss(\vv_t)
                -\frac{\eta}{4}\|\grad\loss(\pw_t)\|_2^2 
                + \frac{\eta}{2}\|\upperH(\vv_t - \pw_t)\|^2
                + \frac{\eta}{2}\|\grad\loss(\vv_t) - \vb_t\|_2^2 \\
            &\leq \loss(\vv_t)
                -\frac{\eta}{4}\|\grad\loss(\pw_t)\|_2^2 
                + \frac{5\eta}{2}\|\upperH(\vv_t - \pw_t)\|^2
    \end{align*}
    Next, we see
    \begin{align*}
        \pw_{t+1} - \pw_t &= -\eta\grad\loss(\pw_t)-\eta\cz_t\\
        \vv_{t+1}-\vv_t &= -\eta \grad\loss(\pw_t)+\eta\upperH(\pw_t-\vv_t)\\
        \pw_{t+1} - \pw_t - (\vv_{t+1}-\vv_t) &=-\eta\upperH(\pw_t-\vv_t)-\eta\cz_t\\
        \pw_{t+1} - \vv_{t+1} &= (\mI - \eta\upperH)(\pw_{t} - \vv_{t}) - \eta\cz_t \\
                          &= -\eta\sum_{k=0}^t(\mI - \eta\upperH)^k\cz_{t-k}    \\
        \pw_{t} - \vv_{t}     &= -\eta\sum_{k=0}^{t-1}(\mI - \eta\upperH)^{k}\cz_{t-k-1} 
    \end{align*}
    Plug into the above:
    \begin{align*}
        \loss(\vv_{t+1}) 
            &\leq \loss(\vv_t)
                -\frac{\eta}{4}\|\grad\loss(\pw_t)\|_2^2 
                + \frac{5\eta}{2}\|\upperH(\vv_t - \pw_t)\|^2 \\
            &= \loss(\vv_t)
                -\frac{\eta}{4}\|\grad\loss(\pw_t)\|_2^2 
                + \frac{5\eta^2}{2}\|\upperH\sum_{k=0}^{t-1}(\mI - \eta\upperH)^{k}\cz_{t-k-1}\|^2
    \end{align*}
    Rearranging terms to move $\loss(\pw_t)$ to the left,  then summing  $t=0, \dots, T-1$ gets us:
    \begin{align*}
        \frac{\eta}{4} \sum_{t=0}^{T-1}\|\grad\loss(\pw_t)\|_2^2
            &\leq \loss(\vv_0) - \loss(\vv_{T})
                + \frac{5\eta^2}{2} \sum_{t=0}^{T-1}\|\upperH\sum_{k=0}^{t-1}(\mI - \eta\upperH)^{k}\cz_{t-k-1}\|^2    \\
        \frac{1}{T}\sum_{t=0}^{T-1}\|\grad\loss(\pw_t)\|_2^2
            &\leq \frac{4\loss(\vv_0) - 4\loss(\vv_{T})}{\eta T}
                + \frac{10\eta}{T}\sum_{t=0}^{T-1}\|\upperH\sum_{k=0}^{t-1}(\mI - \eta\upperH)^{k}\cz_{t-k-1}\|^2
    \end{align*}
    Let $\mX = \mC^\top\mC$, $\mW = \mm \odot \mV^\top\Mu^2\mV$ where
    \begin{align*}
        \mm = \begin{pmatrix}
            T & T -1  & \cdots & 1 \\
            T-1 & T-1 & \cdots & 1 \\
            \vdots          & \vdots           & \ddots  & \vdots \\
            1 & 1  & \cdots & 1
        \end{pmatrix}/T
    \end{align*}
    Take expectation w.r.t $\mZ$ gets us:
    \begin{align*}
        \frac{1}{T}\sum_{t=0}^{T-1}\E\|\grad\loss(\pw_t)\|_2^2
            &\leq \E\frac{4\loss(\vv_0) - 4\loss(\vv_{T})}{\eta T}
                + {10\eta\sigma^2}\Tr(\mW\mX^{-1})    \\
            &\leq \frac{4\loss(\vw_0) - 4\loss^*}{\eta T}
                + {10\eta\sigma^2}\Tr(\mW\mX^{-1})
    \end{align*}
    That completes the proof.
\end{proofEnd}

\subsection{Selecting the Criterion for the Mixing Matrix}
The agreement between the quadratic case (Theorem \ref{thm:risk}) and the non-convex case (Theorem \ref{thm:dpmf-fosp-Hsmooth}) is a useful sanity check that the intuition built on the quadratic case can carry over to deep networks. However, there still remains the question of whether one should choose a mixing matrix $\mC$ that minimizes the right side of Equation \ref{eqn:thm:q1} (recall $\mX=\mC^\top\mC$) or the right hand side of Equation \ref{eqn:thm:q2} (in both cases, with the $\mM$ and $\mV$ matrices computed from the eigenvalues of a Hessian upper bound). 

One might argue that the way to design theoretically-motivated algorithms is by minimizing 
the upper bound on convergence rate, $\frac{1}{T}\sum_{t=0}^{T-1} \|\grad \loss(\pw_t)\|_2^2$, as this is 
the quantity that most non-convex convergence analyses study. However, 
fast convergence does not always imply the final model is good. Consider the following example:

\begin{example}
    Consider a network where all layers use the ReLU activation function. Consider a pathological private training algorithm that, for each iteration $t$, sets all weights to 0 and sets the biases to be negative values (e.g., $-100$). In such a network, all activation outputs are 0; the loss over real data would be high, but the gradients are all 0. In this setting, the pathological training algorithm achieves the best possible value of $\frac{1}{T}\sum_{t=0}^{T-1} \|\grad \loss(\pw_t)\|_2^2$ (namely, 0), but clearly results in a useless model. 
\end{example}

Thus, while $\frac{1}{T}\sum_{t=0}^{T-1} \|\grad \loss(\pw_t)\|_2^2$ measures the rate of convergence, it is not necessarily related to the quality of the final result. 
For DP training, we are more interested in the final model's quality (loss/accuracy), as the final accuracy not being good is the main issue we are aiming to solve.
Hence, we opt to use Equation \ref{eqn:thm:q1} as our criterion for choosing a good mixing matrix $\mC$.

Our proposed objective for $\mC$ can be computed as follows. Given a $p\times p$ Hessian  upper bound $\upperH$ (Section \ref{sec:upper_bound}) and desired number of private training iterations $\T$, compute the eigenvalues $\mu_0,\dots, \mu_{p-1}$ (including multiplicity) of $\upperH$. Define $\mM$ to be the $p\times p$ diagonal matrix $\text{Diag}(\mu_0,\dots, \mu_{p-1})$. Define $\mV$ to be the $p\times\T$ matrix such that $\mV[i,j]=(1-\eta\mu_i)^{\T-j-1}$. Then the quality of the mixing matrix $\mC$, used for correlating noise, is $$\Tr((\mC^\top\mC)^{-1}\mV^\top\mM\mV).$$

We next explain how to integrate this criterion into the DP-MF framework (adding privacy constraints for optimization and performing the subsequent model training).


\subsection{Integration with DP-MF}
\label{sec:application_to_dp_mf}


%

\ftrl \citep{bandmf} is a state-of-the-art DP-MF framework with privacy amplification via batches. In the version of DP they use, two datasets $D$ and $D'$ are neighbors if one can be obtained from the other by blanking out 1 record. 
\begin{definition}[\citet{approxdp}]
  Given privacy parameters $\epsilon>0$ and $\delta\in[0, 1]$, a mechanism (randomized algorithm) $M$ satisfies $(\epsilon,\delta)$-differential privacy if for all pairs of neighboring datasets $D, D'$ and all subsets $S\subset \text{range}(M)$, $P(M(D)\in S) \leq e^\epsilon P(M(D')\in S) + \delta$.
\end{definition}
Given the privacy parameters $\epsilon, \delta$, and a desired iteration count $\T$, \ftrl optimizes $\mC$ to minimize its objective function under the following constraints for privacy:  (1) each diagonal entry of $\mX = \mC^\transpose \mC$ is 1, and (2) $\mX$ is positive definite and (3) banded
(i.e., $\mX_{i,j} = 0, \forall |i-j| \geq b$, where the band size $b$ is a tunable hyperparameter and allows privacy amplification).
We can simply replace the objective function of \ftrl with ours and leave the rest of the framework unchanged, which gives the following optimization problem:
%
\begin{problem}
    \label{p:primary}
    \begin{array}{rl}
         \mX              &\gets \argmin\Tr((\mV^\transpose\mM\mV)\mX^{-1}) \\
        \text{s.t.} & \diag(\mX) = \one      \\
                          & \mX \succ 0  \\
                          & \mX_{i,j} = 0, \forall |i-j| \geq b    
    \end{array}
\end{problem}
%
As in the original \ftrl~\citep{bandmf},
this convex problem that can be solved by optimizers like L-BFGS~\citep{lbfgs} with projections onto the feasible region to enforce $\diag(\mX) = \one$, and starting with an initial guess for $\mX$ that is symmetric and positive definite. 
Following \ftrl, once an $\mX = \mC^\transpose\mC$ is obtained, one can use the Cholesky decomposition to recover $\mC$ as a lower triangular matrix. 

\begin{algorithm}[!h]
    \caption{\ftrl Training}
    \label{algo:bandmf-train}
    \textbf{Input:} Initial model $f$ with parameters $\vw_0 \in \R^p$, dataset $\dataset$, mixing matrix $\mC \in \R^{T \times T}$, clip norm $\clipnorm$, 
    noise $\vz \in \R^{T \times p}$ with entries $\sim \N(0, \sigma^2)$,
    learning rate $\eta$ 
    \\
    \textbf{Output:} DP trained model $\vw_T$
    \begin{algorithmic}[1]
        \FOR{$t = 1, 2, \dots, T$}
            \STATE{Create a \ftrl batch $B_t \subseteq \dataset$}
            \STATE{$\vg_i = \nabla\ell(f(\vx_i; \vw_t), y_i)$ for $(\vx_i, y_i) \in B_t$}
            \STATE{$\tilde{\vg_i} = \min(1, \clipnorm / \|\vg_i\|_2)\vg_i$}
            \STATE{$\vw_t = \vw_{t-1} - \eta\left( \sum_{i=1}^{|B_t|}{\tilde{\vg_i}} + \clipnorm(\mC^{-1}\vz)_{[t,:]}\right)$}
        \ENDFOR
    \end{algorithmic}
\end{algorithm}

Pseudocode for \ftrl training is shown in~\Cref{algo:bandmf-train}. For completeness, Appendix \ref{app:psedo} discusses batch creation and privacy accounting (or see~\citet{bandmf} for full details). It differs from regular \dpsgd in two ways --- batch creation and noise addition. \textbf{Batch Creation:} given a band hyperparameter $b$ and batch size $|B|$, the dataset $D$ is initially divided into $b$ equal sized partitions $D_1,\dots, D_b$. During model training, the batch $B_t$ for iteration $t$ is obtained by (1) selecting partition $D_\ell$, where $\ell=1+(t \mod b)$ then (2) setting the batch $B_t$ to be a random sample from $D_\ell$ of size $|B|$ (\Cref{algo:bandsampling}). \textbf{Noise Addition:} using $\mC$ to obtain correlated noise $\widetilde{\vz}_0,\dots, \widetilde{\vz}_{\T-1}$ in an online manner is essentially Gaussian elimination (\Cref{app:psedo}).
Our new $\mC$ from Problem~\ref{p:primary} can be directly integrated into \Cref{algo:bandmf-train}'s noise addition step.



\section{Estimating the Hessian Upper Bound}
\label{sec:upper_bound}

Our new objective relies on finding a matrix $\upperH$ that upper-bounds the Hessian of the loss along the training path, i.e., $\upperH \succeq \grad^2\loss(\pw_t)$. In other words, instead of upper bounding the Hessian of the loss at all possible parameter settings of the network, we just need to upper bound it for a ``good' region of the parameter space. In this section, we present a mathematical and empirical analysis which indicates that unlabeled public data can be used for this purpose. The benefit of public data is that there is no privacy cost for using it. We also empirically evaluate the effect of out-of-distribution public data and show that the upper bound estimation is still robust.
\subsection{Estimating the Upper Bound with Pre-Training}  \label{sec:hess-ub}

Our approach for estimating the Hessian upper bound is fairly simple. 
First, we assign random labels to examples in the public dataset and pre-train the model with those random labels. Although this seems counterintuitive, we provide mathematical and empirical support for this approach in  Sections \ref{sec:ub_logreg} and \ref{sec:ub_dnn}. In particular, Section \ref{sec:ub_logreg} builds intuition with a mathematical analysis of logistic regression, while Section \ref{sec:ub_dnn} provides additional empirical support for deep networks.  

After pre-training ends, we estimate the Hessian on the public dataset at the resulting parameter values, compute its eigenspectrum, and set negative eigenvalues to 0 (since we are seeking a Hessian upper bound). For relatively small networks, Hessian computation can be done using autograd. For large networks, we note that the objective for the mixing matrix $\mC$ in Theorem \ref{thm:risk} only depends on the eigen-spectrum of the Hessian. Hence, we use the Lanczos method \cite{lanczos}, which uses efficient Hessian-vector products, to estimate the top-$k$ (e.g., top 1000) eigenvalues and use curve-fitting to estimate an upper bound on the tail. The details are in Section \ref{sec:fit}. 

%

\subsection{Hessian Upper Bounds and Logistic Regression}
\label{sec:ub_logreg}

It seems counterintuitive that a Hessian upper bound can be computed from random labels. After all, where is the signal coming from? To build intuition, we consider a tractable case where the Hessian exists in closed form. Thus, we first study the setting of logistic regression pre-trained on in-distribution data with random labels. We show that this yields a meaningful Hessian upper bound and that random labels are actually preferable to true labels (in case the public data were labeled).

%
Logistic regression is a convex model for binary classification. Given a weight vector $\vw$ and a feature vector $\vx_i$, its prediction is $\sigma(\vw\cdot\vx_i)$, where $\sigma$ is the sigmoid function.
Ordinarily, it is trained using negative log-likelihood loss/binary cross-entropy. That is, if the $y_i$ are the true labels for the $\vx_i$, the loss $\loss(\vw)$ is
\begin{align*}
    -\sum_{i=1}^n[ y_i\log(\sigma(\vx_i^\transpose \vw)) + (1 - y_i)\log(1 - \sigma(\vx_i^\transpose \vw))].
\end{align*}

The true Hessian at $\vw$ is $\grad^2 \loss(\vw_t) = \mX^\top \mD_{\vw} \mX$, where $\mX$ is the matrix whose rows are the feature vectors, and $\mD_{\vw}$ is the diagonal matrix whose $(i,i)$-th entry is $\sigma(\vw\cdot\vx)(1 - \sigma(\vw\cdot\vx))$. It will be important to note that $\sigma(\vw\cdot\vx)(1 - \sigma(\vw\cdot\vx))\leq \frac{1}{4}$.

Now, with random labels, each $y_i$ is equally likely to be 0 or 1 and is unrelated to the feature vector. A logistic regression trained with random labels will try to predict $0.5$ for everything to indicate its uncertainty, and this can be achieved by setting the weight vector to the 0 vector: $\vw=\mathbf{0}$. For this setting, the unsupervised Hessian is $\grad^2 \loss(\mathbf{0}) = \mX^\top (\frac{1}{4}\mI) \mX$.
Now, since  $\mD_{\vw}$ is diagonal with every entry $\leq \frac{1}{4}$, $\frac{1}{4}\mI-\mD$ is positive semi-definite, and therefore,
\begin{align*}
    \grad^2 \loss(\mathbf{0}) - \grad^2 \loss(\vw) &=  \mX^\top (\frac{1}{4}\mI) \mX - \mX^\top \mD_{\vw} \mX
\end{align*}
is also positive semi-definite. In short, the unsupervised Hessian $\grad^2 \loss(\mathbf{0})$ is an upper bound on the true Hessians.

This analysis can provide intuition about why this method should work. First, where is the learning signal for the unsupervised pretraining coming from? Since the Hessian upper bound here is $\grad^2 \loss(\mathbf{0})=\frac{1}{4}\mX^\top\mX$, the learning signal is coming from the data feature vectors, and the eigenspectrum of the Hessian upper bound is determined by the singular values of $\mX$. Hence, unsupervised pre-training learns about characteristics of the data features. Second, are random labels better than the true labels? From the formula above, a supervised Hessian with respect to a weight vector $\vw_t$ is unlikely to be an upper bound for the Hessian at some other parameter vector $\vw_{t^\prime}$. In other words, using the true labels makes the Hessian too specific to the weight vector it is taken at. Finally, why do random labels work? They force the logistic regression to be maximally uncertain (e.g., always predicting 0.5), which places its pre-activations near the most responsive part of the sigmoid (i.e., the middle), rather than at the extremes which would cause saturation.

\subsection{Hessian Upper Bounds and Deep Networks}
\label{sec:ub_dnn}
The intuition from the logistic regression analysis also makes intuitive sense for deep networks: random labels force the network to make predictions with maximum uncertainty, which helps avoid saturation of intermediate activation functions, and hence helps increase the eigenvalues of the Hessian that is computed when pre-training ends. The important signal comes from the feature vectors, such as input images in computer vision tasks. Even when data sets have vastly different meanings, like CIFAR-10 vs. a chest x-ray dataset, there are many similar intermediate tasks that a network must perform (e.g., implicit edge detection) so that this approach may even be robust when the public dataset is out-of-distribution with respect to the private training data. In this section, we empirically evaluate the Hessian upper bound construction under various settings.

%
%
%
%
%

\paragraph{Empirical Upper-Bound Analysis.} 
To what extent does the unsupervised Hessian upper bound $\upperH$, computed from public data, serve as a true upper bound to the Hessians encountered during the private training weight update sequence $\pw_1,\dots, \pw_\T$ and the non-private training weight update sequence $\vw_1,\dots,\vw_\T$? The following theorem introduces a dominance certificate $\rho(\mH, \upperH)$ which quantifies the extent to which $\upperH$ is an upper bound on $\mH$. 

%
%

\begin{theoremEnd}[proof end,category=hess-ub]{theorem}[Dominance Certificate]\label{thm:hess-ub}
    Let $\mH$ and $\upperH$ be two symmetric real matrices, with $\upperH$ positive semi-definite. Define (using the convention 0/0=0):
    \begin{align*}
        \rho(\mH, \upperH) 
            := \max_{\vv \neq \zero}\frac{\vv^\top \mH \vv}{\vv^\top \upperH \vv}.
    \end{align*}
    Then,
        $\mH \preceq \upperH \Longleftrightarrow \rho(\mH, \upperH) \leq 1$.
\end{theoremEnd}

\begin{proofEnd}
   Note that $\rho(\mH,\upperH)\leq 1 \Leftrightarrow \max_{\vv}\vv^\top \upperH \vv - \vv^\top \mH \vv \geq 0$
   because $\vv^T\upperH\vv\geq 0$ since $\upperH$ is positive semi-definite. The right hand side is the definition of a Hessian upper bound.
%
%
%
%
\end{proofEnd}

\begin{figure}[t]
  \centering
  \includegraphics[width=0.49\textwidth]{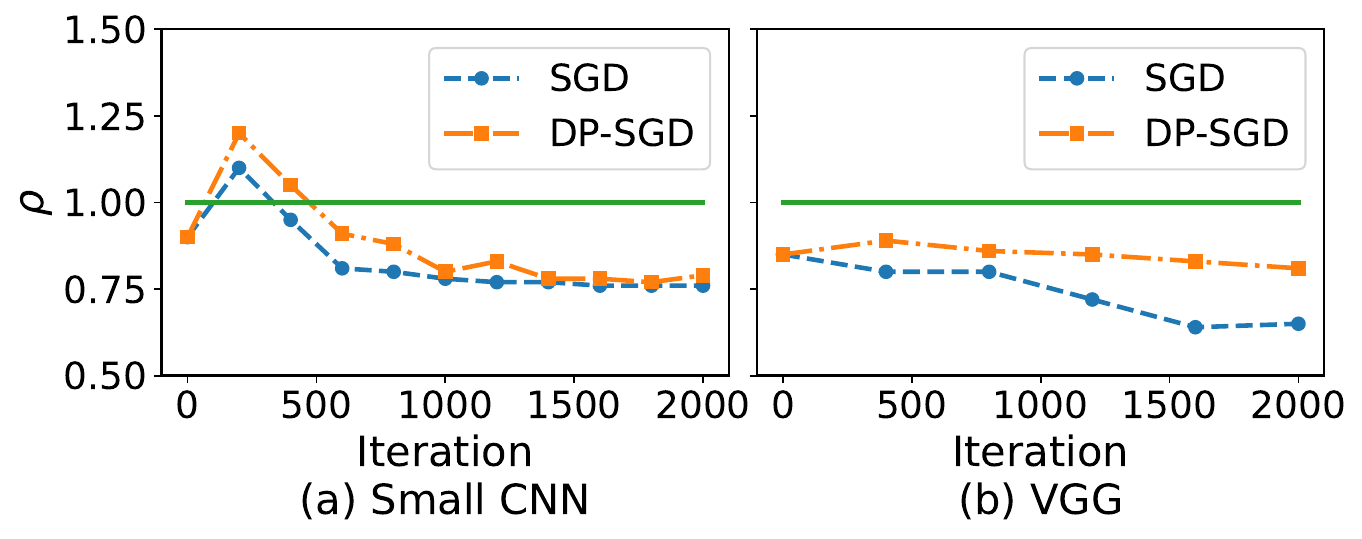}
  \caption{Dominance certificate $\rho$ over training, calculated every (a) 200 and (b) 400 iterations. Training data is CIFAR-10 and public data is TinyImageNet.
  Architecture details are in~\Cref{app:expsetup}.
  }
  \label{fig:hess-ub-rho}
\end{figure}

If  $\rho(\mH, \upperH) \leq 1$, then $\upperH$ is an upper bound. If $\rho>1$, it indicates by how much we ``missed'' the upper bound (i.e., by how much $\upperH$ needs to be rescaled  in order to become an upper bound). 
%
%
%
%
We use this dominance certificate $\rho$ to evaluate the upper bound $\upperH$ that is obtained from public data. 
For the first set of results, the public dataset is TinyImageNet and the private training set is CIFAR-10. We train both a non-private model on CIFAR-10 and also a private model using our proposed method (called \techname). In both cases, we periodically compute the dominance certificate. The models we use include a small CNN (architecture explained in \Cref{app:expsetup}) and a VGG network. The results are shown in Figure \ref{fig:hess-ub-rho}.



\Cref{fig:hess-ub-rho} shows that our estimated $\upperH$ is a good approximate upper bound. For the small model, in some early iterations, $\upperH$ is off by a factor of at most 1.25 but otherwise serves as a true upper bound on the Hessians encountered during private and non-private training. For VGG, $\upperH$ was always a true upper bound.

%
%


\begin{figure}[t]
  \centering
  \includegraphics[width=0.45\textwidth]{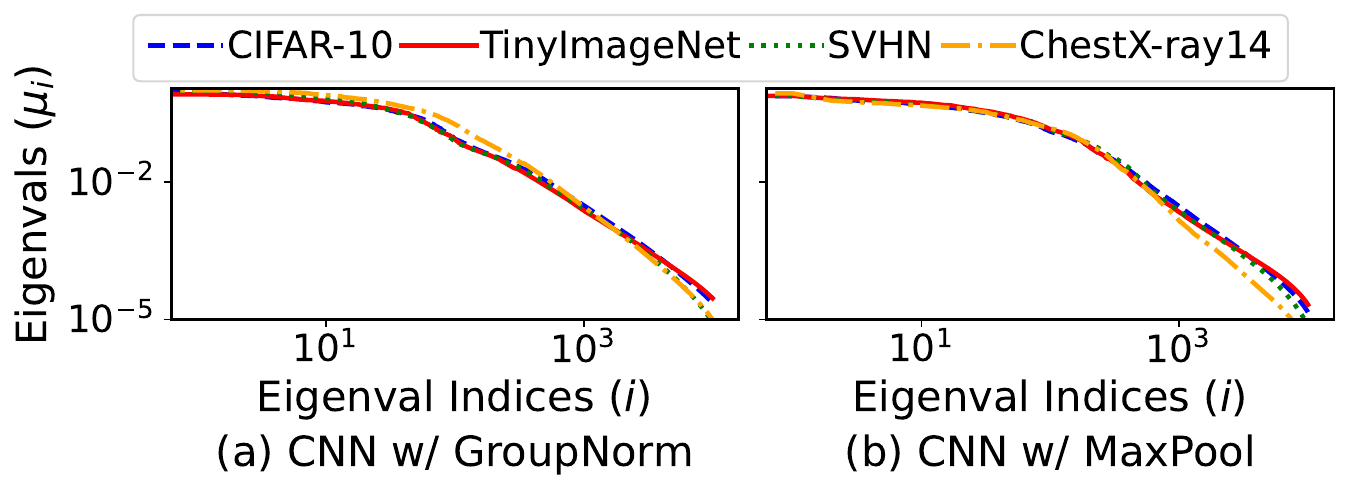}
  \caption{Largest 10,000 eigenvalues of four datasets estimated on two CNNs. Setup details are in~\Cref{app:expsetup}.
  }
  \label{fig:evals-pub}
\end{figure}

\paragraph{Impact of Public Dataset Similarity.}
The next question is: how robust is the estimation of $\upperH$ to public datasets that are highly dissimilar to the training data? We evaluate this question in Figure \ref{fig:evals-pub}.
Since only the eigenvalues of $\upperH$ affect the choice of the noise correlation mixing matrix $\mC$, we plot the eigenspectrum of an $\upperH$ derived from each of the following 4 datasets:
CIFAR-10~\citep{cifar10}, TinyImageNet~\citep{tinyimagenet}, SVHN~\citep{svhn}, and ChestX-ray14~\citep{chestxray}. If CIFAR-10 is considered to be the private training dataset,  TinyImageNet would be the closest public dataset. SVHN (street view house number photos) is quite different and would be considered out of distribution, and ChestX-ray14 (X-ray images) is the most dissimilar to CIFAR-10. However, the eigenspectra of the unsupervised Hessian upper bounds computed on each dataset are remarkably similar (Figure \ref{fig:evals-pub}). We speculate that the reason is that computer vision datasets require similar implicit processing in the lower layer of a network (e.g., edge detection, corner detection, identifying textures, etc.) and hence the data features that influence unsupervised Hessians tend to be similar.

\subsection{
Upper Bound Estimation for Larger Models
}  \label{sec:fit}

\begin{figure}[t]
  \centering
  \includegraphics[width=0.49\textwidth]{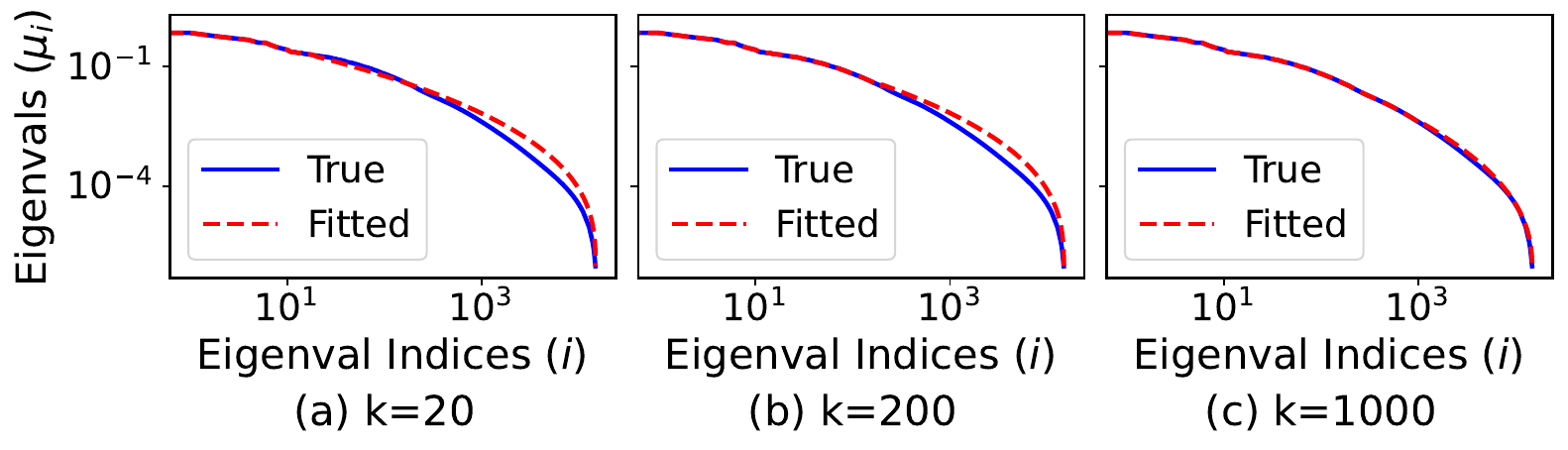}
  \caption{The true and fitted eigenvalues in descending order of a small CNN ($\sim$30,000 parameters), using the top $k = 20, 200, 1000$ eigenvalues for fitting (normalized by the largest). 
  Details in~\Cref{app:expsetup}.
  }
  \label{fig:fit}
\end{figure}

For very large models, direct calculation of $\upperH$ requires significant computational overheads. Our new objective (Problem~\ref{p:primary}) only uses the 
\emph{eigenvalues} of $\upperH$ and not the entire matrix itself, but even so, calculating all the eigenvalues may  still be computationally infeasible.
%
%
%
%
%
%
Our proposed solution is to calculate as many eigenvalues as is feasible, and then fit a curve to upper bound the tail (that is, we don't need to estimate the tail accurately, we just need to upper bound it).
We first compute the $k$ largest positive eigenvalues up to a point allowed by the compute budget, using Lanczos~\citep{lanczos} algorithm. Then we note that eigenvalues of deep learning models tend to follow a power law on the log-log scale (see, for example, Figure~\ref{fig:fit} that plots the eigenspectrum of models for which the entire eigenspectrum can be computed). A log-log power law looks like Equation \ref{eq:fit} 
%
%
\begin{equation}    \label{eq:fit}
    \log \mu_i=C \cdot\left[\log\left(\frac{p_{+}}{i}\right)\right]^\alpha+\log \mu_{p_{+}},
\end{equation}
where log of the $i^\text{th}$ largest eigenvalue $\mu_i$ is equal to a power $\alpha$ of the log of its position $\log(1/i)$. 
The parameters $C$ and $\alpha$ are fitted to the $k$ computed eigenvalues. $\mu_{p_+}$ represents the smallest eigenvalue we would care about (we use $\mu_{p_+}=10^{-6}$ in large model experiments). $p_+$ represents the position of the smallest eigenvalue $\geq \mu_{p_+}$.

We compute it by solving a cumulative empirical spectral measure (CESM) problem.
CESM is a problem of calculating how many eigenvalues of a matrix lie below a threshold. 
We use stochastic Lanczos quadrature (SLQ) with Hessian-vector products, following \citet{slq}, to solve this problem and thereby calculate $p_+$. As shown in Figure \ref{fig:fit}, the fitted curves serve as a good upper bound on the tail of the eigenspectrum empirically. 

%
Specifically, the fitted curve matches the true curve for the first $k$ eigenvalues, but the two curves diverge afterwards (eigenvalues beyond $k$ are the tail). The fitted curve remains an upper bound on the tail of the eigenspectrum -- the log-log scale of the plots makes the difference appear to be visually small, so the gap is larger than it appears.

\section{Evaluation Results}\label{sec:experiments}

We compare our proposed \techname to \ftrl and \dpsgd under various privacy budgets $\epsilon$, with $\delta=10^{-5}$. 
For vision tasks, we use CIFAR-10~\citep{cifar10} and ChestX-ray14~\citep{chestxray} datasets for training, and trained/finetuned a small CNN, ResNet~\cite{resnet}, VGG~\cite{vgg}, and ViT~\cite{vit}. 
For all vision tasks, we use TinyImageNet as the public dataset for pre-training.
For the language task, we finetune a pretrained RoBERTa~\citep{roberta} on MNLI~\citep{glue} using LoRA~\citep{lora}, and use WikiText as the public data.
We split 10\% of data samples from training sets as validation sets and \textbf{tune hyperparameters on validation sets} (no tuning on the test sets).
Further details on our experimental setup can be found in~\Cref{app:expsetup}.

%
%
 
We first note that our results for \ftrl differ from those reported by \citet{bandmf}. We could only match those results if we allowed \ftrl to use the test set for hyperparameter tuning. 
As this is also a common practice in the DP literature, we additionally show the result of hyperparameter tuning on the test set in 
\Cref{app:tune-on-test}, where \techname still consistently outperforms competitors. 

\subsection{Results for Vision Tasks}  \label{sec:exp-large}

\begin{table}[t]
    \caption{Test mean accuracy and standard deviation (over 6 runs).
    }
    \subfloat[Last layer finetuning of ResNet152 (convex loss)]{
        \begin{adjustbox}{width=0.47\textwidth}
        \begin{tabular}{c|c|ccc}
        \hline
        Non-private            & $\epsilon$ & \dpsgd & \ftrl & \techname  \\ \hline
        \multirow{3}{*}{82.01} & 1          & 75.89 $\pm$ 0.01      & 75.95 $\pm$ 0.04     & \textbf{76.46 $\pm$ 0.23} \\
                               & 2          & 77.14 $\pm$ 0.17      & 77.33 $\pm$ 0.09     & \textbf{78.3 $\pm$ 0.05}  \\
                               & 5          & 78.31 $\pm$ 0.23      & 78.35 $\pm$ 0.07     & \textbf{80.14 $\pm$ 0.28} \\ \hline
        \end{tabular}
        \end{adjustbox}
    }
    \hfill
    \subfloat[Full finetuning of a small CNN (nonconvex loss)]{
        \begin{adjustbox}{width=0.47\textwidth}
        \begin{tabular}{c|c|ccc}
        \hline
        Non-private            & $\epsilon$ & \dpsgd & \ftrl & \techname  \\ \hline
        \multirow{3}{*}{72.99} & 1          & 48.67 $\pm$ 0.26      & 49.17 $\pm$ 0.22     & \textbf{49.8 $\pm$ 0.25}  \\
                               & 2          & 51.82 $\pm$ 0.3       & 52.15 $\pm$ 0.64     & \textbf{54.11 $\pm$ 0.42} \\
                               & 5          & 54.64 $\pm$ 0.76      & 55.39 $\pm$ 0.66     & \textbf{58.08 $\pm$ 0.17} \\ \hline
        \end{tabular}
        \label{tab:small-cnn}
        \end{adjustbox}
    }
    
    \label{tab:small-results}
\end{table}

\paragraph{Small Models}
We compare our method with \dpsgd and \ftrl on CIFAR-10 classification with two models: (1) finetuning the last layer of an ImageNet-pretrained ResNet152, where the loss is convex, (2) fully finetuning a small CNN, whose loss is nonconvex.
%
For both, we estimate the full $\upperH$ with the method from Section~\ref{sec:ub_dnn}.

Table~\ref{tab:small-results} summarizes the result: \techname consistently outperformed the baselines in all setups we studied, and \textit{this result is statistically significant}. Specifically, for any setting of $\epsilon$ and architecture, each of the 3 runs of \techname had higher accuracy than each of the runs of \ftrl and DP-SGD. Statistical significance can be assessed using a permutation test: 
if we mix the 6 runs from each row (3 from \techname and 3 from \ftrl) and re-assign 3 each to \techname and \ftrl, there are $6!/(3!3!)$ ways of re-assigning.
%
However, there is only one assignment that makes all 3 runs of \techname to have higher accuracies than \ftrl (which is what we observe). 
Hence, the p-value is $3!3!/6!=0.05$ for every $\epsilon$ setting and every architecture (ResNet and CNN). Put another way, every row in Table~\ref{tab:small-results} is statistically significant.

\paragraph{Larger Models}
For large models where the full $\upperH$ is infeasible to compute, we used curve fitting (Section~\ref{sec:fit}) with \techname to upper bound the tail of the eigenspectrum.

Figure \ref{fig:vgg} (in Section \ref{sec:intro}) compares our proposed \techname to \ftrl and DP-SGD on the VGG training setup used by \citet{multiepoch-mf} on CIFAR-10. The results were averaged over 6 runs and were evaluated for $\epsilon=1,2,5,8$. At $\epsilon=1$, the differences were not statistically significant. However, for $\epsilon=2$, each of the 6 runs of \techname had higher accuracy than all runs of \ftrl. Under the permutation test, this is highly statistically significant with $p$-value$=\frac{6!6!}{12!}\approx 0.001$ (i.e., there are $\frac{12!}{6!6!}$ ways of ordering 6 runs of \techname and 6 runs of \ftrl but there is only one ordering in which all the \techname runs are the largest ones). Similarly, each run of \techname outperformed all runs of \ftrl for $\epsilon=5$ and $\epsilon=8$, with p-value$\approx 0.001$ in both cases.

We also experimented with finetuning ViT with LoRA on ChestX-ray14 (\Cref{tab:cxr}).
%
The ChestX-ray14 experiment represents a case when the public data (TinyImageNet; natural images) does not closely align with the target data (ChestX-ray14; medical images). The results are shown in \Cref{tab:cxr}. For $\epsilon=1,2$ all methods had AUC near 0.5 (i.e., they were no better than random guessing). However, all runs of \techname at $\epsilon=5$ outperformed all runs of the competitors (similarly, p-value $\approx 0.001$ under permutation testing) and same with $\epsilon=8$ (also p-value $\approx 0.001$ under permutation testing). The results show that, even under a significant domain shift (TinyImageNet vs. ChestX-ray14), \techname was able to improve the accuracy over the baselines.


We additionally tried using the validation set of CIFAR-10 and ChestX-ray14 as the public dataset, which represents cases when a public dataset highly similar to the target data is available.
While the accuracy improved with better public data as expected, the additional improvement was only around 0.1--0.14\% for CIFAR-10 and was a moderate 1.47--1.62\% for ChestX-ray14. The result implies that a close dataset like TinyImageNet for CIFAR-10 is good enough as a public dataset. TinyImageNet also serves as a reasonable public dataset for ChestX-ray14, although the accuracy could further improve with an even similar public data.

%
%

\begin{table}[t]  
\caption{
Test mean AUC and standard deviation (over 6 runs) on the  ChestX-ray14 dataset (ViT, LoRA finetuning). for $\delta=10^{-5}$ as $\epsilon$ varies. $\epsilon$=1--2 result omitted as the AUC was too low for all methods (50\% AUC is random guessing).
}
\centering
\begin{adjustbox}{width=0.49\textwidth}
\begin{tabular}{c|c|c|c|c}
\hline
Non-private            & $\epsilon$ & \dpsgd & \ftrl & \techname  \\ \hline
\multirow{2}{*}{73.63} & 5          & 58.16 $\pm$ 0.62      & 59.32 $\pm$ 0.27     & \textbf{62.55 $\pm$ 0.5}  \\
                       & 8          & 59.21 $\pm$ 0.29      & 62.55 $\pm$ 0.43     & \textbf{64.52 $\pm$ 0.69} \\ \hline
\end{tabular}
\end{adjustbox}
\label{tab:cxr}
\end{table}



\subsection{Results for a Language Task}  \label{sec:exp-text}

\begin{table}[t]  
\caption{Test mean accuracy and standard deviation (over 6 runs) on the MNLI dataset (Roberta, LoRA finetuning).}
\centering
\begin{adjustbox}{width=0.49\textwidth}
\begin{tabular}{c|c|ccc}
\hline
Non-private            & $\epsilon$ & \dpsgd & \ftrl & \techname \\ \hline
\multirow{3}{*}{83.73} & 2          & 46.17 ± 0.93          & 50.72 ± 0.41         & \textbf{53.87 ± 0.66}    \\
                       & 5          & 54.73 ± 2.64          & 63.8 ± 1.37          & \textbf{67.18 ± 0.55}    \\
                       & 8          & 63.77 ± 0.7           & 69.57 ± 0.82         & \textbf{73.41 ± 0.82}    \\ \hline
\end{tabular}
\end{adjustbox}
\label{tab:mnli}
\end{table}

\Cref{tab:mnli} summarizes our results for the language task, MNLI. This is another setting where there is a
significant domain shift between the private data (MNLI) and the public dataset (WikiText).
As in vision tasks, \emph{\techname consistently outperforms both baselines}, showing that \techname benefits not only vision models but also language models. 
Again, each of the three runs of \techname at $\epsilon=2$ scored higher than all of the competitor runs for a $p$-value of $\frac{6!6!}{12!}\approx 0.001$ under permutation testing, and the same goes for $\epsilon=5$ and $\epsilon=8$. Hence, all of the results are statistically significant.

\section{Discussion and Limitations}
We propose \techname, which improves the quality of cross-iteration noise correlation through model curvature information. 
Although \techname achieves non-trivial accuracy improvements over SOTA baselines, we recognize a few limitations that require further study.

1. \techname needs an unlabeled public dataset. However, assuming such public data is common in the field~\cite{unlocking, nasr2023effectively}, and our results showed that \techname still benefits from not-so-close datasets (e.g., TinyImageNet vs. ChestX-ray14).

2. \techname requires estimating the eigenvalues of the Hessian of the loss, whose computational complexity grows with the model size. While our curve fitting approximation and engineering alleviate the complexity to some extent, the computational overhead currently prohibits applying \techname when training very-large (e.g., billion-parameter) models. Future algorithmic or hardware-level innovations may address the limitation.

\section*{Acknowledgments}
This work was supported by the US National Science Foundation under Awards CNS-2317232 and CNS-2349610.
Any opinions, findings, and conclusions or recommendations expressed in this material are those of the author(s) and do not necessarily reflect the views of the National Science Foundation.

\bibliography{cite}
\bibliographystyle{icml2026/icml2026}

\newpage
\appendix
\onecolumn

\section{Experiment Details and Results}    \label{app:exp}
We use PyTorch~\citep{pytorch} as the deep learning framework and datasets and models provided by its \texttt{torchvision} package (if exists) or HuggingFace. We implement DP-MF training as a new optimizer under Opacus~\citep{opacus}, a library that enables training PyTorch models with differential privacy, which is also used for our privacy accounting. Our implementation, especially the online noise vector generation, is highly inspired by~\citet{apple-pfl}.

All the experiments in this paper are runnable on one single NVIDIA A5000 24GB GPU.
\subsection{Experiments Setups}    \label{app:expsetup}
\paragraph{Datasets} \textbf{Vision tasks:} We use a variety of datasets, including SVHN~\citep{svhn}, CIFAR-10~\citep{cifar10} as two canonical vision tasks. We also evaluate our methods on one medical image dataset, ChestX-ray14~\citep{chestxray}, as closer to privacy-sensitive applications.
When using ChestX-ray14, we perform a multi-label classification, where each image may be annotated with up to 14 thoracic disease categories (e.g., pneumonia, effusion, cardiomegaly, mass). Due to its class imbalance, we compute Area Under the ROC Curve (AUC) for each class and use the average of them (mean AUC) as the accuracy metric.
Unless otherwise stated, we use 6,000 TinyImageNet~\citep{tinyimagenet} data samples as the public dataset to pretrain the models (\Cref{fig:vgg,fig:evals-pub,fig:fit}, and the main results). We also use those data samples to estimate Hessian/eigenvalues of Hessian for our~\Cref{p:primary} unless specified otherwise. 
\Cref{fig:fit} used CIFAR-10 validation set for the Hessian estimation.
\textbf{Language tasks:} We use MNLI~\citep{glue} as the dataset for language tasks, which is a large-scale benchmark dataset for natural language inference. Each example contains a premise sentence and a hypothesis sentence, labeled as entailment, contradiction, or neutral, i.e., a 3-class classification task. To improve fine-tuning efficiency, we split 10\% of its full data samples for training, and we superisely found that its noise-free accuracy is close to fine-tuning on 100\% data samples as reported in~\citet{dpft-llm}. For the public dataset, we use WikiText~\citep{wikitext}, which is a collection of large-scale language-modeling datasets built from high-quality Wikipedia articles, designed to reflect natural, long-range text rather than heavily preprocessed sentences.

\paragraph{Network Architectures} \textbf{Vision tasks:} Our evaluation use both CNN-based (e.g., ResNet152~\citep{resnet}, VGG~\citep{vgg}) and Transformer-based (ViTs~\citep{vit}) architectures.
For some experiments, we use small-scale CNNs (\Cref{tab:small-cnn-arch}), whose full Hessian can be computed on our computers. When using a small CNN, except for~\Cref{fig:evals-pub} where we use both, we use CNN with GroupNorm (\Cref{tab:cnn-gn-arch}), i.e., in \Cref{fig:fit} and \Cref{tab:small-cnn}. For ResNet152 and ViT (\texttt{vit\_b\_16}), we use the models (together with pretrained weights) provided by \texttt{torchvision}. For VGG, we use the network architecture from~\citet{multiepoch-mf}, which is smaller than a standard VGG.
\textbf{Language tasks:} For the classification task on MNLI, we use a pretrained RoBERTa model (\texttt{ROBERTA\_BASE}) provided by \texttt{torchtext}.

\begin{table}[!h]
\caption{Network architectures of two small-scale CNN models.}
\centering
\subfloat[CNN with GroupNorm]{
    \begin{adjustbox}{width=0.47\textwidth}
    \begin{tabular}{ll}
        \hline
        \textbf{Layer} & \textbf{Parameters} \\
        \hline
        Conv2d + GroupNorm + ReLU & $3 \to 16$, kernel $3\times3$, stride $1$ \\
        Conv2d + GroupNorm + ReLU & $16 \to 32$, kernel $3\times3$, stride $2$ \\
        Conv2d + GroupNorm + ReLU & $32 \to 64$, kernel $3\times3$, stride $2$ \\
        AdaptiveAvgPool2d & output $2\times2$ \\
        Linear + ReLU & $256 \to \texttt{dense\_size}$ \\
        Linear & $\texttt{dense\_size} \to \texttt{num\_classes}$ \\
        \hline
    \end{tabular}
    \label{tab:cnn-gn-arch}
    \end{adjustbox}
}
\subfloat[CNN with Maxpool]{
    \begin{adjustbox}{width=0.53\textwidth}
    \begin{tabular}{ll}
        \hline
        \textbf{Layer} & \textbf{Parameters} \\
        \hline
        Conv2d + ReLU + MaxPool2d & $3 \to 16$, kernel $3\times3$, stride $1$; pool $2\times2$ \\
        Conv2d + ReLU + MaxPool2d & $16 \to 32$, kernel $3\times3$, stride $2$; pool $2\times2$ \\
        Conv2d + ReLU & $32 \to 64$, kernel $3\times3$, stride $2$ \\
        AdaptiveAvgPool2d & output $2\times2$ \\
        Linear + ReLU & $256 \to \texttt{dense\_size}$ \\
        Linear & $\texttt{dense\_size} \to \texttt{num\_classes}$ \\
        \hline
    \end{tabular}
    \end{adjustbox}
}
\label{tab:small-cnn-arch}
\end{table}

\paragraph{Training/Finetuning Strategy} We use full parameter training (for small CNNs and VGG), parameter-efficient fine-tuning mechanisms including freezing layer (for finetuning the last layer of ResNet152) and LoRA~\citep{lora} (for ViT). For fine-tuning ViTs with LoRA, we use low rank $r=8$, making the total number of finetuning parameters around 1M. For fine-tuning RoBERTa with LoRA, we use low rank $r=16$, making the total number of finetuning parameters around 3M. Based on the recommendations of~\citet{choose}, we finetune those parameters on public data first, then perform privacy-preserving training/finetuning on the private target datasets. 

\paragraph{Hyperparameters} Unless otherwise stated, we split 10\% of data samples from training sets as validation sets and \textbf{tune hyperparameters on validation sets}. All models are trained for 20 epochs on CIFAR-10 with batch size 450, 5 epochs on ChestX-ray with batch size 150, and 3 epochs on MNLI with batch size 50. We use constant learning rate, no momentum and no weight decay. We use Optuna~\citep{optuna} for auto hyperparameter tuning and search within $[0.1, 1.]$ for the learning rate and $[0.1, 10.]$ for clip norm. Additionally, for \ftrl and \techname, we search within $[1, 20]$ for the band size.

\subsection{What if Hyperparameters are tuned on Test set (a common practice in the DP literature)} \label{app:tune-on-test}
We re-evaluate the experiments in~\Cref{fig:vgg} by training the model on the \textbf{full training set} and \textbf{tuning hyperparameters on the test set}. Under this setup, we are able to reproduce results comparable to those reported in~\citet{bandmf}, and our method even achieves higher accuracy. Nevertheless, we emphasize that this is a bad practice and should not be adopted for fair comparison—even in noise-free settings.

\begin{table}[!h]  
\caption{Test accuracy comparison on CIFAR-10 under various $\epsilon$ ($\delta=10^{-5}$). We use the \textbf{full} train set for training data and tune hyperparameters on test set. Other setups follow~\Cref{fig:vgg}.}
\centering
\begin{tabular}{c|c|ccc}
\hline
Non-private            & $\epsilon$ & \dpsgd & \ftrl & Our            \\ \hline
\multirow{4}{*}{82.68} & 1          & 54.68                 & 54.68                & \textbf{55.99} \\
                       & 2          & 59.01                 & 59.01                & \textbf{61.12} \\
                       & 5          & 63.39                 & 65.59                & \textbf{68.2}  \\
                       & 8          & 65.06                 & 67.19                & \textbf{72.01} \\ \hline
\end{tabular}
\label{tab:tune-on-test}
\end{table}

\subsection{Statistical Significance Tests}  \label{sec:sig-test}
We performed both statistical significance tests (one-sided paired t-test with H1: ``our method better than DP-BANDMF", and the permutation test) for each choices of $\epsilon$ in~\Cref{fig:vgg}. For paired t-test, its $t$-statistic and $p$-value can be computed using \texttt{scipy} package, and we calculate the $p$-value of permutation test directly. 
\begin{itemize}
    \item $\epsilon=1$: $t = 1.69, p = 0.076$  (not significant -- the differences are too small to be detectable with 6 seeds).
    \item $\epsilon=2$: $t=8.3, p < 0.001$ for the one-sided paired t-test. For the permutation test of comparing NoiseCurve (6 seeds) to DP-BandMF (6 seeds), there are $12! / (6!6!) = 924$
     orderings of their results. Thus, under the null hypothesis of no difference between the two methods, the probability that every run of NoiseCurve would have the higher accuracy is 1/924 for a p-value of approximately $p=0.001$.
     \item $\epsilon=5$: $t=6.32, p < 0.001$ for the one-sided paired t-test. The same analysis as for the $\epsilon=2$ case yields $p=0.001$ for the permutation test.
     \item $\epsilon=8$: $t=19.49, p < 0.00001$  for the one-sided paired t-test. The same analysis as for the $\epsilon=2$ case yields $p=0.001$ for the permutation test.
\end{itemize}

For the rest of the experiments (\Cref{tab:small-results,tab:cxr,tab:mnli}, which run over 3 seeds for each cell), we observed that NoiseCurve consistently gets higher accuracy than \ftrl, meaning that the permutation test yields a $p = 1 / (6! (3! 3!)) = 0.05$.

\subsection{In-distribution Public Data}    \label{app:in-dist-pub}
In addition to using TinyImageNet as public data for eigenvalue estimation, we also evaluate accuracy when eigenvalues are estimated from the validation set of the target dataset, which can be regarded as in-distribution data relative to the training set.

\begin{table}[!h]
    \caption{Comparison of test accuracy of using TinyImagenet vs. Validation set as public data under various $\epsilon$ ($\delta=10^{-5}$). CIFAR-10 experiments following settings of~\Cref{fig:vgg} and ChestX-ray experiments follow~\Cref{tab:cxr}.}
    \centering

    \subfloat[CIFAR-10]{
        \begin{adjustbox}{width=0.45\textwidth}
        \begin{tabular}{c|c|cc|cc}
        \hline
        Non-private            & $\epsilon$ & \dpsgd & \ftrl  & Tiny  & Val  \\ \hline
        \multirow{2}{*}{84.05} & 5          & 62.17  & 63.29 & 65.77 &  65.91    \\
                               & 8          & 64     & 64.46 & 69.91 & 70.1 \\ \hline
        \end{tabular}
        \end{adjustbox}
    }
    \subfloat[ChestX-ray14]{
        \begin{adjustbox}{width=0.45\textwidth}
        \begin{tabular}{c|c|cc|cc}
        \hline
        Non-private            & $\epsilon$ & \dpsgd & \ftrl & Tiny  & Val   \\ \hline
        \multirow{2}{*}{73.63} & 5          & 58.39                 & 59.71                & 62.67 & 64.14 \\
                               & 8          & 59.66                 & 61.99                & 64.28 & 65.9  \\ \hline
        \end{tabular}
        \end{adjustbox}
    }

    \label{tab:indist}
\end{table}

 We observe that using validation data eigenvalues yields little-to-no improvement on CIFAR-10 and a modest gain of 1–2\% on ChestX-ray14. These results indicate the robustness of our method: even when only out-of-distribution public data such as TinyImageNet are available, the accuracy improvement remains.

\subsection{What if not Pretrain}   \label{app:not-pretrain}

We further evaluate the setting without pretraining, where eigenvalues are estimated under the default initialization of the deep learning framework. 

\begin{table}[!h]  
\caption{A comparison of test accuracy of VGG on CIFAR-10 with vs. without pretrain. Other setups follow~\Cref{fig:vgg}.}
\centering
\begin{tabular}{c|c|ccc}
\hline
$\epsilon$         &              & \dpsgd & \ftrl  & \techname  \\ \hline
\multirow{2}{*}{5} & w/o pretrain & 62.21 & 63.18 & 64.41 \\
                   & w/ pretrain  & 62.17 & 63.29 & 65.77 \\ \hline
\multirow{2}{*}{8} & w/o pretrain & 64.71 & 65.89 & 68.15 \\
                   & w/ pretrain  & 64    & 64.46 & 69.91 \\ \hline
\end{tabular}
\label{tab:pretrain-or-not}
\end{table}

\begin{figure}[!h]
    \centering
    \includegraphics[width=0.5\linewidth]{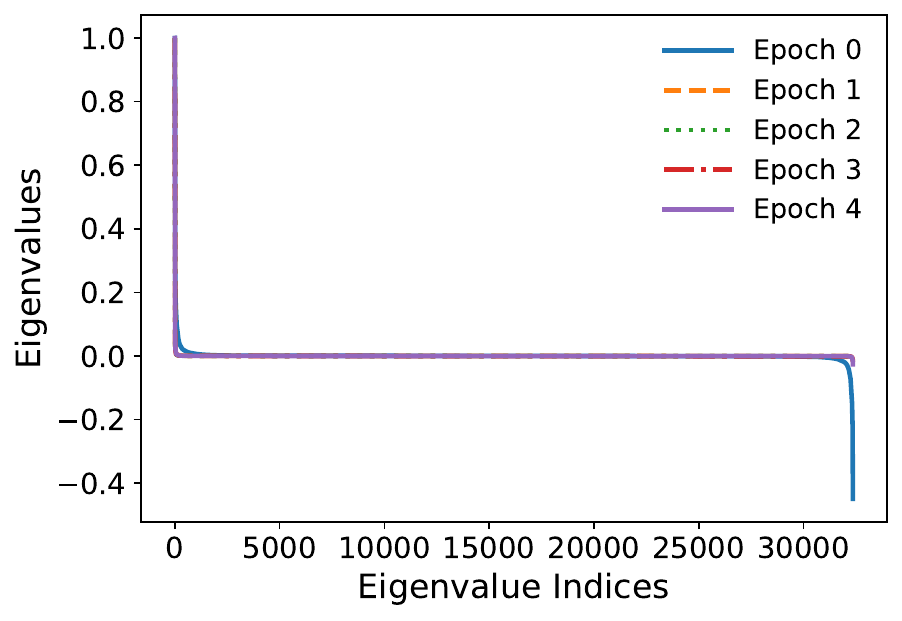}
    \caption{Change of the Hessian eigenvalues during training. Despite significant negative eigenvalues at initialization, positive eigenvalues dominate.}
    \label{fig:evals-default-init}
\end{figure}

The results show that even without pretraining, our method achieves a 1–2\% accuracy improvement over \dpsgd and \ftrl. We conjecture that this is because modern initialization schemes (e.g., He~\citep{he-init} or Xavier~\citep{xavier-init}) are designed to preserve the variance of activations and gradients across layers, which helps maintain stable signal propagation at the start of training and mitigates the risk of the parameters being placed in regions with pathological curvature (see~\Cref{fig:evals-default-init}). Nevertheless, pretraining remains advantageous: it stabilizes the Hessian spectrum, yielding a nearly constant curvature that strengthens the effectiveness of our method and thereby leads to larger accuracy improvements.


\section{Proofs}\label{app:proofs}
\printProofs[quadratic]
\printProofs[nonconvex]
\printProofs[hess-ub]

\section{Algorithm Pseudo Code} \label{app:psedo}
We present the full pseudocode of the DP-MF training algorithm, adapted from~\citet{bandmf}. To ensure completeness, we additionally present~\Cref{algo:priv-acct} to show how to do privacy accounting under DP-MF. Our method, \techname, modifies it by provide a different the noise correlation $\mC$ as shown in~\Cref{algo:bandmf-train}.

\begin{algorithm}[!h]
    \caption{\ftrl Privacy Accounting}
    \label{algo:priv-acct}
    \textbf{Input:} Privacy budget $(\epsilon, \delta)$, Privacy accountant $\texttt{PRVAccountant}$, Clip norm $\clipnorm$, Batch size $B$, Total number of data samples $n$, Training number of epochs $E$, Number of bands $b$ \\
    \textbf{Output:} Noise scale $\sigma$
    \begin{algorithmic}[1]
        \STATE {Total number of training steps: $\T \gets \frac{EB}{n}$}
        \STATE {Sample rate for \ftrl: $q \gets \frac{Bb}{n}$}
        \STATE {Total number of compositions for \ftrl: $T \gets \frac{\T}{b}$}
        \STATE {Compute noise multiplier: $C \gets \texttt{PRVAccountant}(\epsilon, \delta, q, T)$}
        \STATE {Noise scale: $\sigma \gets C \cdot \clipnorm$}
    \end{algorithmic}
\end{algorithm}

\begin{multicols}{2}
    \begin{algorithm}[H]
        \caption{\ftrl Data Batch Sampling}
        \label{algo:bandsampling}
        \textbf{Input:} Dataset $\dataset$, Batch size $|B|$, Current step $t$ \\
        \textbf{Output:} Data batch $B_t$
        \begin{algorithmic}[1]
            \STATE{$D_1, \dots, D_b$ $\gets$ arbitrary partition of $\dataset$}
            \STATE{$j = t (\text{mod } b)$} 
            \STATE{$B_t \gets$ random size $|B|$ subset of $D_j$}
        \end{algorithmic}
    \end{algorithm}

    \columnbreak
    
    \begin{algorithm}[H]
    \caption{\ftrl Noise Generation Online}
    \label{algo:online-noise}
    \textbf{Input:} Noise correlation $\mC \in \R^{T \times T}$ ($b$-banded and lower triangluar), Noise scale $\sigma$, Current step $t$ \\
    \textbf{Output:} $(\mC^{-1}\mZ)_{[t,:]}$
    \begin{algorithmic}[1]
        \STATE{$\vz \sim \N(0, \sigma^2\mI_p)$}
        \FOR{$i = 1, 2, \dots, T$}
            \STATE{$\vx_i = (\vz_i - \sum_{j=i-b+1}^{i-1}{\mC_{i,j}\vx_j})/\mC_{i,i}$}
        \ENDFOR
        \STATE {\textbf{Return } $\vx$}
    \end{algorithmic}
\end{algorithm}
\end{multicols}


Note that in \Cref{algo:online-noise}, ~\citet{bandmf} simplify the presentation by using the convention that out-of-bounds indexing into a matrix or vector returns 0.





\end{document}